\documentclass[sigconf]{acmart}

\RequirePackage{OpenProc-Mods}

\AddToHook{cmd/maketitle/before}{%

\def\EDBTISSN{2367-2005}
\def\EDBTISBN{978-3-89318-107-0}
\setcopyright{none}
\copyrightyear{\copyright\ 2026 Copyright held by the owner/author(s).
  Published on \url{OpenProceedings.org} under ISBN \EDBTISBN, series ISSN \EDBTISSN.
  Distribution of this paper is permitted under the terms of the
  Creative Commons license CC-by-nc-nd 4.0}
\acmYear{2026}
\acmDOI{}
\acmISBN{}

\acmConference[EDBT '27]{Extending Database Technology}{6-9 April 2027}{Lille (France)}

\settopmatter{printacmref=false, printccs=false, printfolios=false}

\newsavebox{\ximagebox}
\newlength{\ximageheight}
\newsavebox{\xglyphbox}
\newlength{\xglyphheight}
\newcommand{\xbox}[1]%
  {\savebox{\ximagebox}{#1}%
  \settoheight{\ximageheight}{\usebox{\ximagebox}}%
  \savebox{\xglyphbox}{\color{white}\char32}%
  \settoheight{\xglyphheight}{\usebox{\xglyphbox}}%
  \raisebox{\ximageheight}[0pt][0pt]{\raisebox{-\xglyphheight}[0pt][0pt]{%
    \makebox[0pt][l]{\usebox{\xglyphbox}}}}%
    \usebox{\ximagebox}%
    \raisebox{0pt}[0pt][0pt]{\makebox[0pt][r]{\usebox{\xglyphbox}}}}

\newsavebox{\LogoBox}
\sbox{\LogoBox}{\includegraphics[height=1cm]{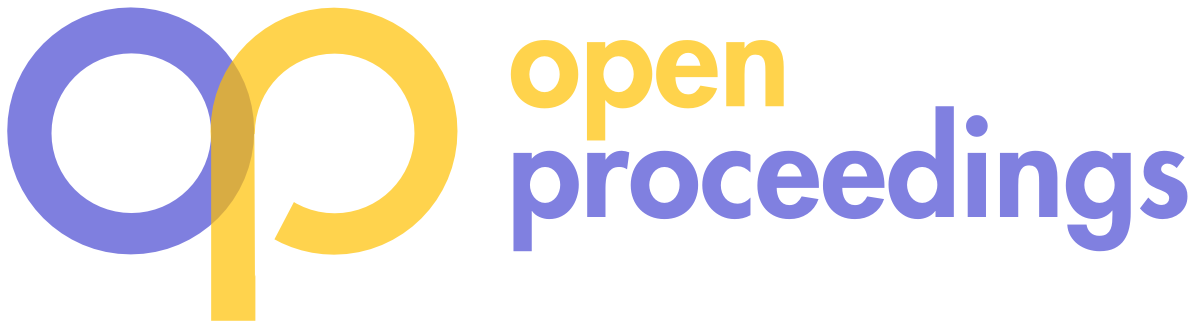}}

\fancypagestyle{firstpagestyle}{%
  \fancyhf{}%
  \fancyfoot{}%
  \fancyhead[L,C]{}
  \fancyhead[R]{\href{https://OpenProceedings.org/}{\raisebox{15pt}[0pt][0pt]{\xbox{\usebox{\LogoBox}}}}}
  }
  
\hypersetup{%
  pdfcopyright={CC-by-nc-nd},
  pdfpublisher={OpenProceedings.org -- University of Konstanz, Germany},
  pdfcreator={LaTeX with acmart, hyperref, and EDBT modifications},
  pdfvolumenum={30},
  pdfissuenum={2},
  pdfisbn={\EDBTISBN},
  pdfissn={\EDBTISSN},
  pdfdoi={10.48786/edbt.2027.xx}
}
}%

\newcommand{\sys}{\textsc{Splice}\xspace}
\newcommand{\sysdp}{\textsc{Splice-DP}\xspace}
\newcommand{\sysgrad}{\textsc{Splice-Grad}\xspace}
\newcommand{\target}{\mathcal{T}}
\newcommand{\sources}{\mathcal{S}}
\newcommand{\profit}{\mathcal{P}}
\newcommand{\gain}{\Delta}
\newcommand{\santos}{\textsc{Santos}\xspace}
\newcommand{\allsrc}{\textsc{AllSrc}\xspace}
\newcommand{\single}{\textsc{Single}\xspace}
\newcommand{\grasp}{\textsc{Grasp}\xspace}
\newcommand{\greedy}{\textsc{Greedy}\xspace}
\newcommand{\random}{\textsc{Random}\xspace}
\usepackage[ruled,vlined,linesnumbered]{algorithm2e}
\usepackage{algorithmic}
\usepackage{enumitem}
\usepackage[dvipsnames]{xcolor}
\usepackage{subcaption}
\usepackage{multirow}
\usepackage{threeparttable}
\usepackage{tabularx}
\usepackage[normalem]{ulem}

\begin{document}

\OPtitle{Scalable Source Selection for Machine Learning Tasks}

\OPauthor[0009-0006-7463-1421]{Ambarish Singh}
\affiliation{%
  \institution{Purdue University}
  \city{West Lafayette}
  \state{IN}
  \country{USA}
}
\email{sing1104@purdue.edu}

\OPauthor[0009-0003-1538-0472]{Romila Pradhan}
\affiliation{%
  \institution{Purdue University}
  \city{West Lafayette}
  \state{IN}
  \country{USA}
}
\email{rpradhan@purdue.edu}

\renewcommand{\shortauthors}{Singh et al.}

\begin{OPabstract}
Given a query table, existing data lake discovery techniques 
seek to identify data sources related to the query table through schema similarity and joinability. The discovered sources, however, are not always useful for downstream analyses; while some 
improve task generalization, some have the potential to inadvertently introduce harmful noise or distribution shifts.
We formalize source selection as the problem of finding a subset of the discovered sources that maximizes downstream task utility. We propose \sys, a target- and cost-aware source selection system that adopts the idea of gene splicing to identify sources which
provide additional signals that the union of all sources
misses.
To speed up the process, we develop two surrogate variants, \sysdp and \sysgrad, that estimate the marginal contribution of subsets to task utility by leveraging data profiles in a meta-learning approach and gradient alignment, respectively. 
On classification tasks over two real-world 
benchmarks derived from SOTA table union search tools,
\sys shows $3-5$ percentage points improvement in binary F1 scores compared to the case when  all discovered tables are considered. 
On five additional real-world ML tasks spanning classification and regression, \sys improves task utility over the all-sources baseline by 1.8–13.7 points, while \sysdp and \sysgrad match this quality with 3–7x fewer model evaluations (up to 77x at scale).  
\end{OPabstract}


\maketitle

\section{Introduction}
\label{sec:intro}
Data lakes have emerged as a powerful storage paradigm to store huge amounts of heterogeneous data extracted from several data sources without the need to enforce a unified schema. With the rapid adoption of machine learning (ML) across domains, ensuring high-quality data is extracted from large data lakes is a top priority in data-driven decision-making systems. 
Given a downstream ML task, preparing its training dataset from an existing data lake primarily consists of two distinct steps: (i)~\emph{discovering} data sources that are relevant to the task, and (ii)~\emph{selecting} discovered sources that are useful to the downstream task utility. 

Source discovery involves identifying candidate sources for a task and is fundamentally a combinatorial search problem over a large corpus~\cite{paton2023dataset, zhu2016lsh}; 
source selection, on the other hand, is a statistical decision problem~\cite{engstrom2024dsdm}
over a much smaller set that involves determining which subset of the candidate sources actually improves downstream task utility. 
Existing tools such as \textsc{Santos}~\cite{khatiwada2023santos}, \textsc{Josie}~\cite{zhu2019josie}, and \textsc{Aurum}~\cite{fernandez2018aurum} focus on efficiently discovering tables from data lakes through techniques such as measuring schema overlap and computing the joinability of the query table and tables in the data lake. However, a data lake table that is joinable with the query table may not necessarily be \emph{useful} for downstream tasks. In other words, a data source that satisfies the task requirements is not guaranteed to result in the best task utility. While data discovery systems often accurately identify sources (e.g., tables) that are suitable for potentially training a model, they are not equipped with the tools to evaluate whether the characteristics of an identified dataset, such as label distribution, feature coverage, or domain shift, are expected to facilitate or impede model training.
We describe the challenges of dealing with these issues in the following example.
\begin{example}

Consider the \textsc{Santos} Large benchmark~\cite{khatiwada2023santos} consisting of $11$k tables collected from Canada and UK Open Data as the data lake. Given a query table 
with schema \texttt{<Department Family}\texttt{, Entity, Date}\texttt{, ExpenseType, ExpenseArea}, \texttt{Suppl-} \texttt{ier}\texttt{, Transaction\#, Amount, VATRegistration\#>}, table union search system \textsc{Santos} returns the top $20$ unionable tables: all monthly spending records from the UK Intellectual Property Office sharing a related schema. Using the query table as the test set and the remaining 19 tables as candidate data sources, we consider the binary classification task of predicting if a transaction is high-spend (target variable \texttt{Amount} exceeding the median amount in the query table) or low spend given its \texttt{ExpenseType} and \texttt{ExpenseArea}.
Subsets of these tables exhibit varying utilities: a model trained on all sources has an F1 score of $0.715$ while one trained only on August 2010 data has F1=0.  
The latter, while schema compatible, has far fewer high-spend transactions and a classifier trained on it wrongly predicts every query transaction as low-spend.
In contrast, a carefully selected subset of $7$ tables has F1 score$=0.745$, exceeding the bound of $0.726$ obtained when the query table is used for both training and testing.
\label{example:motivation}
\end{example}

\begin{figure*}[t]{}
    \centering
    \includegraphics[width = \linewidth]{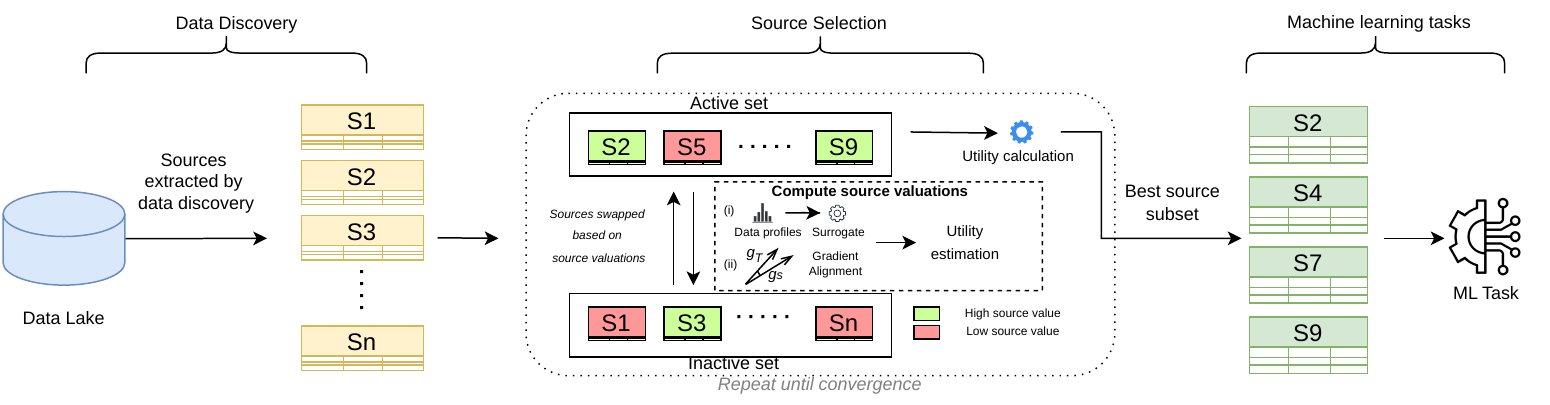}
    \captionof{figure}{An overview of \sys: a gene-splicing-based source selection mechanism that actively adds and removes sources from an active set to output the subset of sources having the best downstream performance.}
    \label{fig:overview}
\end{figure*}

\noindent {\textbf{Considering all sources is not enough}}. Data integration solutions usually consolidate discovered sources into a unified dataset for downstream tasks. However, including every source for the task is not the most optimal choice; more data often leads to generalizability, but is vulnerable to differences in distributions of the query and source data occurring primarily due to three kinds of failures: 
\begin{itemize}[leftmargin=*,topsep=0pt]
\item \textit{Label distribution mismatch} occurs when the distribution of target variable in the query data differs from that of the discovered data sources. For instance, the August 2010 data in Example~\ref{example:motivation} has $36\%$ fewer positive labels than the query data.
\item \textit{Feature coverage gap} is observed when discovered sources and query data differ in the range of observed feature values. For example,  sources from different organizational units or time periods may adopt categorical values missing from the query data, causing vocabulary mismatch in feature space encoding.
\item \textit{Source cancellations} happen when one source has a negative effect on task utility and another improves it such that the effect of the former dominates model behavior and nullifies that of the latter. In Example~\ref{example:motivation}, June 2010 alone reaches an F1 score of 0.736, which the pool of all 19 sources does not match.
\end{itemize}

\noindent The example illustrates a more general challenge: selecting a good subset of sources is not simply a matter of discarding obviously bad ones because the benefit of selecting a source depends on the sources that have already been selected. A source that hurts in isolation may help when combined with a complementary one, and a source that seems individually useful may become redundant - or even harmful - in combination with other selected sources. These inter-source interactions mean that selection cannot be decomposed into evaluating sources one at a time -- a na\"ive strategy adopted by classical greedy approaches. With $n$ candidate sources, the space of possible subsets of sources is $2^n$, and the utility of each subset can only be measured by actually evaluating it on the downstream task. This search space quickly becomes practically infeasible with even a moderate $n$, and the failure modes that make selection necessary in the first place 
prevent simple greedy heuristics from succeeding. 

We leverage the aforementioned insights to propose \sys, an algorithm inspired by gene-based splicing~\cite{berget1977spliced} to efficiently select the subset of data sources best-suited for a downstream ML task. \sys partitions the discovered sources into an \emph{active} and an \emph{inactive} set and iteratively swaps sources that degrade task utility (termed \emph{low value}) for sources that improve it (termed \emph{high value}), converging when the active set no longer changes (Figure~\ref{fig:overview}). Deciding whether a source is of high or low value entails training a model at each iteration and computing its utility - a computationally expensive step for large datasets and complex models. We therefore develop two surrogate variants that estimate the improvement in task utility without retraining: \sysdp, which predicts marginal gains from \emph{data profiles}~\cite{feurer2015efficient} of the current active set using a meta-learning model~\cite{vanschoren2018meta}, and \sysgrad, which ranks candidate subsets by the cosine alignment between their aggregate source gradients and the target gradient (Section~\ref{sec:estimation}).

\noindent Our main contributions can be summarized as follows:
\begin{itemize}[leftmargin=*,topsep=2pt]
    \item Given a set of data sources, we formalize the source selection problem as determining the subset of sources that results in the best utility over a downstream ML task (Section~\ref{sec:prelim}).
    \item We propose \sys, an algorithm based on gene splicing that finds subsets with high utility without exhaustive subset enumeration, and develop two efficient variants, \sysdp and \sysgrad, that respectively use data profiles and {gradient alignment} 
    to estimate the marginal contribution of a subset to utility, substantially reducing the number of model trainings (Sections~\ref{sec:approach} and ~\ref{sec:estimation}).
    \item We empirically demonstrate that \sys and its variants consistently outperform several baselines on five real-world ML tasks and over two \textsc{Santos} benchmarks. Compared to \sys, \sysdp and \sysgrad 
    incur {3–7x savings in model evaluation costs (growing upto 77x as the number of candidate sources increases)} 
    with interactive response times (Section~\ref{sec:exp}).
\end{itemize}

\section{Related Work}
\label{sec:related}
This research is at the intersection of data discovery, data valuation, and subset selection for ML tasks. 
Source selection addresses gaps in these lines of research by determining relevant candidates to include in training a target-specific model.

\noindent\textbf{Data discovery and integration in lakes.} Existing work on data discovery~\cite{cafarella2009data, sarma2012finding, paton2023dataset, nargesian2019data,nargesian2018table,lehmberg2017stitching} emphasize identifying candidate tables in a data lake that join or union with a query e.g., \textsc{Josie}~\cite{zhu2019josie},
\textsc{Santos}~\cite{khatiwada2023santos},
\textsc{Aurum}~\cite{fernandez2018aurum},
\textsc{LSH Ensemble}~\cite{zhu2016lsh},
and D$^3$L~\cite{bogatu2020dataset}.
These systems answer \textit{"which tables in the lake are related to given query ?"},
but not \textit{"which of these related tables improve a downstream task ?"}.  \sys treats discovery tools as black-box upstream oracles and addresses this selection question. (In Section~\ref{sec:santos}, we show that this gap is empirically significant.) 
Different from join-based feature augmentations e.g.,
ARDA~\cite{chepurko2020arda}, METAM~\cite{galhotra2023metam}, \sys selects unionable sources 
under a black-box, non-monotone utility. 
A complementary line of research \textsc{Sloth}~\cite{zechhini2024sloth} removes redundancy
by finding the largest overlap between tables to detect near-duplicates in data lakes. While \sys automatically assigns near-zero marginal profit to duplicates of active sources, such preprocessing boosts  \sys since its cost at every swap is linear in source pool size.

\noindent\textbf{Data valuation and selection.} 
Data Shapley~\cite{pmlr-v97-ghorbani19c} and influence functions~\cite{pmlr-v70-koh17a} value training data at example granularity,
requiring $O(2^{N})$ retrainings for $N$ examples or $O(p^3)$ Hessian inversion for $p$ model parameters in the exact case; even practical approximations require per-example retraining or matrix inversion. \sys differs from data valuation along three axes: it operates at the \textit{granularity of sources} 
which is the natural unit of provenance, licensing, and access control in data lakes~\cite{nargesian2019data}; its objective is a \textit{target-specific}  utility; and it explicitly handles \textit{non-monotone} utility where adding a source can decrease performance -- a regime where Shapley's cooperative game axioms require careful re-interpretation. The closest to our work is model-aware dataset selection~\cite{engstrom2024dsdm} (DsDM), 
but it requires a large number of model trainings to be effective.

\noindent\textbf{Coreset, meta-learning and search heuristics.} Coreset methods select example-level subsets to approximate full-data training via gradient alignment~\cite{killamsetty2021grad}, generalization objective~\cite{killamsetty2021glister}, submodular maximization~\cite{mirzasoleiman2020coresets}, and geometric coverage~\cite{sener2017active}. \sysgrad adapts the gradient-alignment principle from example to source granularity.
Selection using coresets compresses a pool already deemed useful.
Unlike coresets that assume a fixed distribution and unified training set, we operate over heterogeneous sources with an explicit target; Section~\ref{sec:coresets} shows that point-level coresets cannot avoid whole-source harm. \sysdp surrogate adapts meta-features used by AutoML systems~\cite{rivolli2022meta, vanschoren2018meta, feurer2015efficient, drori2021alphad3m} to predict marginal \textit{utility gains} from profile \textit{differences} between subsets.
Classical subset-search metaheuristics e.g., genetic algorithms~\cite{tan2008genetic}, simulated annealing~\cite{kirkpatrick1983optimization}, tabu search~\cite{glover1990tabu}, and \grasp~\cite{bermejo2011grasp} treat the subset space as black-box optimization requiring multiple utility evaluations per candidate. \sys's swaps are closest to these methods in structure, but our surrogate estimators bypass re-evaluation cost.

\section{Preliminaries} %
\label{sec:prelim}

\noindent\textbf{Data sources.} 
Let $\sources = \{S_1, \ldots, S_n\}$ denote a pool of candidate sources where every source $S_i = \{(\mathbf{x}_{i,j}, y_{i,j})\}_{j=1}^{m_i}$ is assumed to be a labeled collection. 
Let $\target = \{(\mathbf{x}_k^{(T)}, y_k^{(T)})\}_{k=1}^{m_T}$ be a labeled target on the same feature representation as the sources. 
We assume that all sources and the target have been preprocessed through preprocessing steps, such as schema matching, label identification, and feature transformation, to share the feature representation required for the downstream task.
Following standard practice in validation-guided data selection~\cite{engstrom2024dsdm, pmlr-v97-ghorbani19c, killamsetty2021grad, pmlr-v70-koh17a}, the target $\target$ represents a small labeled sample of the target distribution that guides selection; the model trained on the selected sources is intended for future, unseen data from that distribution. We write $D(S) = \bigcup_{S_i \in S} S_i$ for the union of any subset $S \subseteq \sources$.
We assume access to a set of data sources each of which is a collection of examples that share schema required for the downstream task. Sources may differ in their label distributions, feature coverage, or generative process in general.
In a data lake, sources are typically tables returned by a discovery tool such as \textsc{Santos}~\cite{khatiwada2023santos} that joins or unions with a query table. More generally, a source can be any cohesive grouping of labeled examples e.g., records from a U.S. state in a census dataset, images from camera-trap locations, or reviews of a single product category. Our problem formulation and solutions treat sources as opaque, set-valued units, and our framework  applies regardless of data modality.

\vspace{1mm}\noindent\textbf{Machine learning task.} 
We consider supervised learning tasks where the goal is to train a model $\mathcal{M}: \textsf{Dom}(\mathbf{X}) \rightarrow \textsf{Dom}(Y)$ on a dataset $\textsf{D}$ such that each data point $\mathbf{x} \in \mathbf{X}$ has an associated predicted output $\hat{y} = \mathcal{M}(\mathbf{x}) \in \textsf{Dom}(Y)$. The output space $\textsf{Dom}(Y)$ may be binary ($\{0, 1\}$), categorical ($\{1, \ldots, c\}$), or real-valued ($\mathbb{R}$), corresponding to binary classification, multi-class classification, and regression respectively. We use $h_S$ to denote a model of class $\mathcal{M}$ trained on $D(S)$ using a fixed learning algorithm and evaluated on $\target$ under a fixed task utility $\mathcal{U}$. 
We instantiate the model with  logistic regression, linear regression, or a shallow head (linear or small MLP) over frozen features (details in Section~\ref{sec:setup}).

\vspace{1mm}\noindent\textbf{Profit score.}
We summarize model quality on $\target$ by a real-valued \emph{profit} function $\profit_\target(S) \in \mathbb{R}$ that scores the subset $S$. 
Depending on the task, $\profit_\target(S)$ could be a single performance metric (e.g., binary F1 score, accuracy, mean-squared error) or a combination of metrics (e.g., a weighted average of accuracy and true positive rate parity).

\noindent\textbf{Problem statement.}
Given these preliminaries, we define the \emph{source selection problem} below.
\begin{definition}[Budgeted Source Selection]
Given the set of candidate sources $\mathcal{S}$, query table $\mathcal{T}$, profit function $\mathcal{P}$, and budget $B \leq |\mathcal{S}|$,
we seek to determine the best subset of sources that maximizes profit. Formally, 
\begin{equation}
S^{*} \;=\; \arg\max_{S \subseteq \mathcal{S},\; |S| \le B}\ \mathcal{P}_\mathcal{T}(S).
\label{eq:source-selection}
\end{equation}
\label{def:source-selection}
\end{definition}

\noindent \textbf{Hardness of source selection.} The search space for Eq.~\ref{eq:source-selection} 
comprises $2^{|\mathcal{S}|}$ subsets rendering the problem NP-hard (Theorem ~\ref{thm:nphard}). 

\begin{theorem}[NP-hardness of Source Selection]
\label{thm:nphard}
The budgeted source selection problem (Definition~\ref{def:source-selection}) is NP-hard, even when $\profit$ is a monotone set function and the budget
$B=k$ is fixed.
\end{theorem}

\begin{proof}[Proof sketch]
We reduce from \textsc{Max-$k$-Coverage}, which is NP-hard \cite{karp2009reducibility}. An instance of \textsc{Max-$k$-Coverage} consists of a universe $U = \{e_1,\dots,e_m\}$, a collection of sets $\mathcal{F} = \{F_1,\dots,F_n\}$ with each $F_i \subseteq U$, and an integer $k$. The goal is to find $k$ sets whose union covers the maximum number of elements of $U$.
Given such an instance, construct a source selection instance as follows. The target $\target$ consists of $m$ labeled examples $\{(e_j, 1)\}_{j=1}^m$, one per universe element. Source $S_i$ consists of the labeled examples $\{(e_j, 1) : e_j \in F_i\}$. Define the profit function $\profit_\target(S)$ as the number of target examples covered by the union of sources in $S$, i.e. $\profit_\target(S) = |\bigcup_{S_i \in S} S_i|$. 
A solution to source selection with budget $|S| \le k$ achieving $\profit_\target(S^\ast) = \mathrm{OPT}$ corresponds exactly to a solution to \textsc{Max-$k$ Coverage} covering $\mathrm{OPT}$ universe elements. 
The reduction is polynomial. Since \textsc{Max-$k$-Coverage} is NP-hard, so is source selection.
\end{proof}

\begin{corollary}
Unless P~$=$~NP, no polynomial-time algorithm approximates the budgeted source selection problem within a factor better than $1 - 1/e \approx 0.632$ \cite{feige1998threshold}.
\end{corollary}
{Note that the stated bound applies to the monotone special case; the general (non-monotone) setting we consider in this paper admits no closed-form approximation guarantee.}

\noindent \textbf{Greedy selection.} 
The cost of any source selection strategy is dominated by the number of times it must train and validate a model on some subset of sources. The na\"ive greedy solution builds a subset up to size $k$ by repeatedly adding the best single source from $n$ sources and has $n\cdot k$ utility evaluations,
leading to an $O(n\cdot k \cdot T_{\mathrm{train}})$ complexity where $T_{\mathrm{train}}$ is the cost of a utility evaluation by training the model on a candidate subset and scoring it on the target. This strategy is infeasible for large $n$ and complex models.

\section{\sys Algorithm}
\label{sec:approach}
We now describe our algorithm for selecting the best subset of sources that has a high profit score for a given task. We go through an example scenario, provide key insights, and describe algorithm details. We evaluate marginal utilities of individual sources to active set in an iteration, rank sources according to computed marginal utilities, and swap sources between active and inactive sets to form a subset with higher utility than before. The key insight of our approach is to capture the interactions between different sources to evaluate their marginal utilities,  prioritize state of the current active set, and determine the subset of sources with best utility. 

\subsection{Example scenario}\label{sec:example}
We describe the working of \sys on the IPO classification scenario of Example~\ref{example:motivation}, evaluated empirically in Section~\ref{sec:santos}. \sys draws on the idea of gene splicing - a concept originally used for the synthesis of alternate proteins by editing genetic messages (discarding suboptimal non-coding regions and stitching together coding regions)~\cite{berget1977spliced}. Gene splicing has been hugely successful in domains such as healthcare, medicine, statistics and agriculture~\cite{lin2020base, lee2016therapeutic, zhu2020polynomial, li2016gene}, but its application to data management problems has not been well-explored. Adapting splicing to our setting, \sys starts with an initial active set of sources, iteratively swaps out low-value sources in the active set for high-value sources in the inactive set (i.e., sources that do not belong to the active set) until no further improvement in task utility is found, and returns the active set upon convergence. It repeats this process for a range of initial active sets of increasing size (top-$i$ sources by individual profit) and returns the best subset across all runs. By swapping multiple sources at once rather than adding one at a time, \sys avoids the locally optimal subsets that greedy forward selection settles on.

\noindent\textbf{(Step 1)}
\sys computes individual profit scores (i.e., F1 score) of each of the 19 tables, runs over a range of initial subset sizes $i = 1,\ldots,B$ (multi-start loop), and forms active set $A$ from top-$i$ tables ranked by those scores; the remaining tables form the inactive set $\mathcal{I} = \sources \setminus A$. At $i=5$, $A =$ \{Jun 2010, Oct 2010, May 2011, Mar 2011, May 2010\} are the five months with the highest individual F1 on the April 2010 target, and $\mathcal{I}$ holds the remaining 14 months.
 
\noindent\textbf{(Step 2)}
Within each multi-start run, \sys evaluates the contribution of each source $s \in \sources$ to the current active set by computing its \textit{marginal profit}. We denote by $A'$ the candidate set formed by adding (removing) one or
more sources to (from) $A$.

\begin{definition}[Marginal profit {of a source}]
Given the active set $A \subseteq \sources$ and a candidate set $A' \subseteq \sources$, the marginal profit of replacing $A$ with $A'$ (i.e., $A \to A'$) is given by:
$\gain(A, A') = \profit_\target(A') - \profit_\target(A)$
\label{eq:marginal}
\end{definition}

\noindent Two cases are of interest: (a)~for $s \in A$, $\gain^-(s) = \gain(A, A \setminus \{s\})$ is change in utility when $s$ is removed; $\gain^-(s) > 0$ implies utility would \textit{increase} without $s$; (b)~for $s' \in \mathcal{I}$, $\gain^+(s') = \gain(A, A \cup \{s'\})$ is  change in utility when $s'$ is added and $\gain^+(s') > 0$ implies utility would \textit{increase} by including $s'$.
At $i=5$, active set has profit $\profit_\target(A) = 0.738$. \sys computes $\gain^-(s)$ for each $s \in A$ and $\gain^+(u)$ for each $u \in \mathcal{I}$. $\gain^-(\text{May 2010}) \approx +0.003$ marks it as a candidate for removal, while $\gain^-(\text{Jun 2010})$ $\approx -0.005$ implies Jun 2010 should be retained. Among inactive sources, $\gain^+(\text{Apr 2011})$ is largest positive marginal profit.
 
\vspace{0.5mm}\noindent\textbf{(Step 3)}
\sys then swaps $k$ sources between the two sets, following the heuristic of replacing  $k$ sources in $A$ with highest $\gain^-$ with  $k$ sources in $\mathcal{I}$ with  highest $\gain^+$, and executes the swap if it improves target utility. At $k=1$, \sys removes May 2010 and adds Apr 2011, giving $A_{new}$= \{\text{Jun 2010}, \text{Oct 2010}, \text{May 2011}, \text{Mar 2011}, \text{Apr 2011}\} with $\profit_\target(A_{new}) = 0.743$, up from $0.738$. The swap is accepted as new best within current \textsc{Splicing} call, and the process repeats for each swap size $k = 1,\ldots,k_{\max}$, returning best result across all $k$.
 
\vspace{0.5mm}\noindent\textbf{(Step 4)}
\sys repeats Steps 2 and 3 until no swap of any size improves utility and active set converges, then restarts from a new initial active set. The best subset across all multi-start iterations is returned: here $\{$Jun 2010, May 2011, May 2010, Jul 2010, Aug 2010, Sep 2011, Dec 2010$\}$ with $\profit_\target = 0.745$, found at $i=7$ and exceeding the all-sources baseline of Example~\ref{example:motivation} (full results in Section~\ref{sec:santos}).

\subsection{Algorithm details}
\label{sec:algo}
We present the pseudocode of \sys in Algorithm~\ref{alg:splice}. \sys takes the source pool, budget $B$, and maximum swap size $k_{\max}$, and returns the subset with highest profit on the downstream task. It is organized as three nested routines: main routine (lines 1--7) computes individual source profits once, then seeds active set with top-$i$ sources for each initialization size $i = 1,\ldots,B$ and keeps the best result. For each seed, \textsc{FixedSupport} (lines 8--14) calls \textsc{Splicing} repeatedly, capping swap size at $\min(k_{\max}, |A|)$, and exits when returned subset stops changing i.e., when no swap of size up to $k_{\max}$ improves profit. \textsc{Splicing} (lines 15--29) performs the swaps themselves: given an active set $A$ and a cap $K$, it sweeps $k = 1,\ldots,K$, each $k$ starting afresh from $A$, and operates in three phases.

\begin{algorithm}[t]
\SetKwFunction{FMain}{Main}
\SetKwFunction{FFixedSupport}{FixedSupport}
\SetKwFunction{FSplicing}{Splicing}
\SetKwFunction{FProfit}{getProfit}
\SetKwFunction{FEstimate}{EstimateGain}

\caption{\sys: Source selection algorithm}
\label{alg:splice}
\small
\KwIn{Sources $\sources$, Budget $B$, Swaps $k_{\max}$}
\KwOut{Subset $A^*$ with maximum profit $P^*$}
\Indp
    $A^* \gets \emptyset, P^* \gets -\infty$\;
    $\mathcal{L} \gets$ \textsc{SortDescending}($\sources$, \textsc{getProfit}$(\{s\})$)\;
    \For{$i \gets 1$ \KwTo $B$}{
        {
            $A_{\mathrm{init}} \gets \mathcal{L}[1:i]$\;
        }
        $A_{\mathrm{res}}, P_{\mathrm{res}} \gets$ \textsc{FixedSupport}$(A_{\mathrm{init}}, \sources, k_{\max})$\;
        \If{$P_{\mathrm{res}} > P^*$}{
            $P^* \gets P_{\mathrm{res}}, A^* \gets A_{\mathrm{res}}$\;
        }
    }
\Indm

\BlankLine

\textbf{Function} \textsc{FixedSupport}({$A, \sources, k_{\max}$})

\Indp
    $A_{\mathrm{curr}} \gets A,\; A_{\mathrm{prev}} \gets \emptyset$\;
    \While{$A_{\mathrm{curr}} \neq A_{\mathrm{prev}}$}{
        $A_{\mathrm{prev}} \gets A_{\mathrm{curr}}$\;
        $k \gets \min(k_{\max}, |A_{\mathrm{prev}}|)$\;
        $A_{\mathrm{curr}}, P_{\mathrm{curr}} \gets \textsc{Splicing}(A_{\mathrm{prev}}, \sources, k)$\;
    }
    \Return $A_{\mathrm{curr}}, P_{\mathrm{curr}}$\;
\Indm

\BlankLine

\textbf{Function} \textsc{Splicing}({$A, \sources, K$})

\Indp
    $P_{\max} \gets$ \textsc{getProfit} $(A), A_{\mathrm{best}} \gets A, \mathcal{I} \gets \sources \setminus A$\;
    \For{$k \gets 1$ \KwTo $K$}{
        \For{each $s \in A$}{
            $\gain^-(s) \gets $ \textsc{EstimateGain} $(A, A \setminus \{s\})$\;
        }
        $L_{\mathrm{rm}} \gets$ \textsc{SortDescending}($A$, $\gain^-$)\;%
        $A_{\mathrm{temp}} \gets A \setminus L_{\mathrm{rm}}[1:k]$\;
        
        \For{each $u \in \mathcal{I}$}{
            $\gain^+(u) \gets $ \textsc{EstimateGain} $(A_{\mathrm{temp}}, A_{\mathrm{temp}} \cup \{u\})$\;
        }
        $L_{\mathrm{add}} \gets$ \textsc{SortDescending}($\mathcal{I}$, $\gain^+$)\;%
        $A_{\mathrm{new}} \gets A_{\mathrm{temp}} \cup L_{\mathrm{add}}[1:k]$\;
        
        $P_{\mathrm{curr}} \gets$ \textsc{getProfit} $(A_{\mathrm{new}})$\;
        \If{$P_{\mathrm{curr}} > P_{\max}$}{
            $P_{\max} \gets P_{\mathrm{curr}}, A_{\mathrm{best}} \gets A_{\mathrm{new}}$\;
        }
    }
    \Return $A_{\mathrm{best}}, P_{\max}$\;
\Indm
\end{algorithm}

\begin{itemize}[leftmargin=*,topsep=0pt]
\item \textit{Phase 1} (lines 18--21). Marginal profits $\gain^-(s)$ are computed for every $s \in A$ via \textsc{EstimateGain}; the $k$ sources whose removal helps most are dropped, yielding $A_\mathrm{temp}$.
 
\item \textit{Phase 2} (lines 22--25). Marginal profits $\gain^+(u)$ are computed for every $u \in \mathcal{I}$ \emph{against $A_\mathrm{temp}$} rather than against the original $A$; the $k$ best are added, yielding the candidate
$A_\mathrm{new}$.
 
\item \textit{Phase 3} (lines 26--28). $A_\mathrm{new}$ is scored with \textsc{getProfit} and accepted if it beats the best profit seen in this call. Acceptance uses the exact profit even when \textsc{EstimateGain} is a surrogate (Section~\ref{sec:estimation}), so accepted swaps are genuine improvements.
\end{itemize}

\noindent\textbf{Where the budget enters.} The multistart loop seeds the active set at size $i$ (line 4), and every \textsc{Splicing} call removes exactly as many sources as it adds (lines 21, 25), so $|A|=i$ is invariant within that run. The returned $A^*$ is the best over $i=1,\dots,B$
as shown in Definition \ref{def:source-selection}. Tightening $B$ respects the practitioner's constraint and reduces the number of initializations searched.

\noindent
\textbf{On local-optimum guarantee.}
Theorem A.2 establishes that subset returned by \textsc{FixedSupport} is a $k_{\max}$-local optimum of $\mathcal{P}_\mathcal{T}$ (Definition A.1) only when \textsc{Splicing} evaluates all $\binom{|S|}{k}\binom{n-|S|}{k}$ candidate $k$-swaps exhaustively. The implementation  ranks sources by marginal gain and evaluates a single proposed swap per $k$ (lines 20--26), reducing per-step cost from $O\!\left(\binom{n}{k}^2\right)$ to $O(k_{\max} \cdot n)$ at the price of that guarantee. Termination is unaffected---  bound of Theorem A.2 on number of accepted swaps depends on $\mathcal{P}_\mathcal{T}$ being bounded and each accepted swap improving profit by at least $\varepsilon$; practically the algorithm converges, but to a set that need not be $k_{\max}$-locally optimal. Appendix A.3 examines quality of  returned subsets empirically.

\noindent\textbf{Complexity.}
Each iteration of subroutine \textsc{Splicing} computes 
$n$ marginal utility evaluations and one  profit evaluation on  resulting candidate set, leading to a complexity of $O((n+1) \cdot k_{\max} \cdot T_{\text{train}})$ in wall-clock cost.
Each run of  \textsc{FixedSupport} subroutine repeats \textsc{Splicing} until convergence. Theorem A.2 (see Appendix) bounds the number of accepted iterations by 
$T_{\text{conv}} = (\profit_{\max} - \profit_{\min})/\varepsilon$,
leading to 
$O(T_{\text{conv}} \cdot (n+1)\cdot k_{\max} \cdot T_{\text{train}})$ complexity.
Multi-start loop runs \textsc{FixedSupport} $B$ times, leading to an overall complexity of $O(B \cdot T_{\text{conv}} \cdot k_{\max} \cdot (n + 1) \cdot T_{\text{train}})$. {We provide convergence guarantees  in Appendix A.1, and theoretical analysis and empirical results on quality of returned subset in Appendix A.2 and A.3.}
In practice, \sys takes $47$s$-108$ minutes per benchmark on a single core for  pools of $15-30$ sources ($T_{\text{train}}$ takes a few milliseconds to seconds).  Utility estimation with $1000$ sources needs two orders of magnitude more time. To expedite the process, we propose two variants of \sys 
that avoid expensive per-iteration model training, leading to $3$-$7{\times}$ reduction in model evaluations across several benchmarks.

\section{Marginal Utility Estimation}
\label{sec:estimation}

The \textsc{EstimateGain} primitive in Algorithm~\ref{alg:splice} determines the cost of one search step. With exact retraining, each call requires training a model on the candidate subset -- expensive for large datasets or complex models. We develop two surrogates that approximate marginal utilities at a fraction of the cost, drawing on two complementary lines of research. Our first solution \sysdp is based on meta-learning where the goal is
to learn from prior experience across tasks or datasets to predict
performance on new tasks~\cite{vanschoren2018meta} while the second solution \sysgrad is based on the idea of gradient matching~\cite{killamsetty2021grad,mirzasoleiman2020coresets, killamsetty2021glister}.  
These variants follow the same high-level pattern of leveraging precomputed summaries of source and target data to evaluate candidate swaps without extensive retraining. The choice of these techniques is guided by their complementary capability of capturing the source-target relationship via data-distributional fit (\sysdp) and  gradient direction (\sysgrad). We detail the methods in Sections~\ref{sec:dp_surrogate} and~\ref{sec:grad_surrogate} and compare them in Section~\ref{sec:efficient}.

\subsection{\sysdp: Estimation using Data Profiles}
\label{sec:dp_surrogate}

A core ingredient of \sysdp comprises \textit{data profiles} (or, meta-features) -- a compact, fixed-dimensional vector of dataset characteristics such as statistical moments, class distribution, or feature correlations~\cite{rivolli2022meta}. Profiles have previously been used to predict algorithm performance~\cite{brazdil2000comparison, pfahringer2000meta}, recommend hyperparameters and warm-start AutoML optimization~\cite{feurer2015efficient, drori2021alphad3m}. \sysdp adapts this idea to source selection: we treat each candidate subset as a "dataset", summarize it via its profile, and train a regression model that maps profile changes to marginal-utility changes. The key idea is to tailor profiles to the relevance of source selection (e.g., distributional fit to the target, class imbalance, feature-target correlations) rather than to algorithm selection or hyperparameter tuning tasks.
\begin{table*}[t]
\centering
\caption{Data profile summary. $d_\phi$ = profile dimension. Features are semicolon-separated groups, with italic labels indicating what each group captures. Surrogate model column refers to regressor used to predict marginal gains from meta-features.
}
\label{tab:profiles}
\small
\setlength{\tabcolsep}{5pt}
\renewcommand{\arraystretch}{1.2}
\begin{tabular}{p{1.4cm} c p{13cm} p{1cm}}
\toprule
Task & $d_\phi$ & Feature groups & Surrogate \\
\midrule
Tabular classification, regression & 28 &
\textit{Size \& feature shape:} dataset size, PCA-95\% dim, first-PC skew, first-PC kurt, per-feature skew \& kurt (min/max/mean/std); \textit{Target alignment:} per-feature target correlations (min/max/mean/std); \textit{Target structure:} class imbalance, base rate, Bayes-error gap (classification), mean, standard deviation (regression)
& Linear Regression \\
\hline
Image classification & 11 &
\textit{Target alignment:} JS divergence, $\ell_2$ domain gap, centroid dist. (all, tail-class), variance ratio vs target; \textit{Class balance \& coverage:} label \& tail-class coverage, log min/mean class counts, head/tail ratio, within-class compactness
& Gradient Boost \\
\hline
Text classification & 15 &
\textit{Size \& balance:} log size, effective $n$, positive class fraction; \textit{Target alignment:} JS divergence vs target, $\ell_2$ centroid distances ($\mu_{\text{global}}, \mu^+, \mu^-$), cosine similarities to target ($\mu_{\text{global}}, \mu^+, \mu^-$), variance ratio vs target; \textit{Class separation:} Fisher separation, $|$sep $-$ target sep$|$, Fisher ratio, within-class compactness
& Gradient Boost \\
\bottomrule
\end{tabular}
\end{table*}
\sysdp models marginal utility as a function of how the distributional statistics of a candidate subset differ from those of the current subset and of the target. Because tabular and unstructured data differ substantially in their natural feature representations, we instantiate \sysdp with task-appropriate profile features and surrogate model architectures, while sharing the same overall workflow: precompute profiles, train a regression model on observed marginal-gain examples and query the surrogate at inference time.

\vspace{1mm}\noindent\textbf{Profile Construction.} 
For each subset of sources $A$, we compute a profile vector $\phi(A) \in \mathbb{R}^{d_\phi}$ that captures dataset characteristics correlated with task utility. Profiles are broadly categorized as: (a)~dataset-specific statistics (e.g., sample size, label distribution), and (b)~task-specific features tailored to data modality and learning objective. Table~\ref{tab:profiles} summarizes meta-features per task family. Tabular features draw on canonical meta-feature taxonomies from meta-learning literature~\cite{rivolli2022meta, pfahringer2000meta}; for fairness-aware tasks, we also consider base rate and Bayes-error gap between protected groups, which are predictive of fairness outcomes~\cite{kamiran2012data}. Image and text profiles emphasize distributional divergence between source and target -- identified as predictive of transfer performance in domain-adaptation~\cite{ben2010theory}. For tabular tasks, profiles are computed on a 1\% random sample of the pooled source data to control construction cost; image and text profiles are computed on compact pre-extracted feature embeddings. Profile vectors are cached per subset to avoid redundant computation across the search. For projection-based features (e.g. first-PC moments, $\ell_2$ domain gap), we fit a single PCA projection over union of candidate sources once per task and compute these features against the fitted basis, ensuring distance and spread features are comparable across $A$, $A'$ and target $\target$.

\vspace{1mm}\noindent\textbf{Meta-Dataset Construction.}
To train a marginal gain predictor, we build a \emph{meta-dataset} of <change in profile due to adding/removing a source, observed marginal gain> pairs by repeatedly sampling ordered pairs of subsets $(A, A')$ that differ by exactly one source ($A' = A \cup \{s\}$ for addition; $A' = A \setminus \{s\}$ for removal). We generate $m$ additions and $m$ removals for each source, leading to $2mn$ meta training records per task (empirically, $m=5$ suffices). For each pair of addition and removal, we record profiles $\phi(A)$ and $\phi(A')$, target profile $\phi(\target)$, and actual marginal gain $\gain(A,A')=\profit_\target(A')-\profit_\target(A)$; only gain computation requires model training on $A'$. The full set of records is partitioned into training and held-out sets, the surrogate model is fit on the training set and validated on the held-out set.
\begin{algorithm}[]
\SetKwFunction{FEstimateDP}{\textsc{EstimateGain\_DP}}
\caption{\textsc{EstimateGain\_DP} (data profiles)} 
\label{alg:estimate-gain-dp}
\small
\KwIn{$A$, $A'$, $\target$, $f_\theta$, $\phi$}
\KwOut{$\widehat{\Delta}(A, A')$}
\Indp%
    $\phi_A \gets \phi(A)$, $\phi_{A'} \gets \phi(A')$, $\phi_\target \gets \phi(\target)$\;
    \uIf{task is tabular}{
        $x_{\mathrm{meta}} \gets \phi_{A'} - \phi_A$\;
    }
    \Else{
        $x_{\mathrm{meta}} \gets \big[\,\phi_A,\; \phi_{A'},\; \phi_{A'} - \phi_A,\; |\phi_A - \phi_\target|,\; |\phi_{A'} - \phi_\target|\,\big]$\;
    }
    $\widehat{\Delta}(A, A') \gets f_\theta(x_{\mathrm{meta}})$\;
    \Return $\widehat{\Delta}(A, A')$\;
\Indm%
\end{algorithm}

\vspace{1mm}\noindent\textbf{Meta-Surrogate Model $f_\theta$.} 
We instantiate the meta surrogate with two variants depending on task structure. For tabular tasks, the meta-feature vector is the profile difference $\phi(A') - \phi(A)$, and a standardized linear regression is fit to predict $\Delta\profit$. The linear choice reflects our observation that marginal gain depends near-linearly on individual profile shifts (e.g. a small change in class imbalance generates a proportional change in classifier accuracy)~\cite{rivolli2022meta}. For image and text tasks, marginal gain depends on more complex non-linear interactions between subset profile, target profile and their difference. We construct a richer meta-feature vector (Algorithm~\ref{alg:estimate-gain-dp}, line 6) that concatenates absolute profile information with profile differences and profile to target alignment terms. A gradient-boosting regressor is fit on this richer vector to predict $\Delta\profit$, accommodating non-linear interactions between profile components.

\vspace{1mm}\noindent\textbf{Complexity.}
Offline cost is dominated by $2mn$ model evaluations required to build meta-dataset i.e., $O(n \cdot T_{\text{train}})$. Online queries cost $O(d_\phi)$ per candidate move (negligible compared to model training).

\subsection{\hspace{-1mm}\sysgrad: Estimation via Gradient Match}
\label{sec:grad_surrogate}
Our second solution is based on gradient matching, originating in the coreset selection literature for efficient deep learning. GradMatch~\cite{killamsetty2021grad} and related methods~\cite{mirzasoleiman2020coresets, killamsetty2021glister} select training subsets whose aggregate gradients align with a reference gradient (typically the gradient over the full training set or a held-out validation set). The intuition is that two subsets with aligned gradients will move the model parameters in similar directions during training, so a small subset whose gradient matches that of the full set will approximate training over the full set. This idea has been generalized in recent gradient-based data attribution methods~\cite{pruthi2020estimating, park2023trak} that use gradient information to attribute model behavior to specific training examples at scale. \sysgrad lifts this principle from the example level to the source level: rather than selecting individual training points, we score entire sources according to the cosine alignment between their aggregate gradient and the target gradient. This approach reduces the number of gradient vectors stored from $O(N_{total})$ to $O(n_{sources})$ -- typically a reduction of three to four orders of magnitude -- and lets us evaluate candidate swaps via lightweight dot products rather than per-iteration retraining.

\vspace{1mm}\noindent\textbf{Reference Model and Gradient Summaries.}
For each task we use a lightweight reference model whose loss aligns with the evaluation objective.
The architecture is task-specific (a linear or logistic head for tabular tasks, a linear head over frozen embeddings for images, a small MLP head for text), we describe the per task choices in Appendix B. Across tasks, the reference model is trained for a small number of epochs on the union of all source data following the warm-start protocol used in gradient-based coreset selection~\cite{killamsetty2021grad}, this ensures gradients are computed at parameters that reflect the full data distribution rather than random initialization. For metric-aware tasks (macro-F1 on images, binary-F1 on text), gradients are computed against a differentiable soft-F1 loss~\cite{eban2017scalable} so the gradient direction matches the evaluation metric rather than cross-entropy.

Given reference model, we precompute target gradient $g_\target$ on held-out validation split of target data and source gradient $g_i$ for each source $S_i$. For fairness-aware tasks, we additionally store accuracy and TPR-parity gradients $g^{\text{acc}}, g^{\text{fair}}$. For image and text tasks, gradients are $\ell_2$ normalized before storage so subset scoring depends only on direction; for tabular tasks, gradients are stored as per-sample averages (divided by $|S_i|$) since linear-model gradient magnitudes are already directly comparable across sources.

\begin{algorithm}[]
\SetKwFunction{FEstimateGrad}{EstimateGain\_Grad}
\caption{\textsc{EstimateGain\_Grad} (gradient matching)}
\label{alg:estimate-gain-grad}
\small
\KwIn{$A$, $A' = A \cup \{S_i\}$, $g_\target$, $\{g_j\}_{j \in A'}$, $\lambda$ (fairness weight)}
\KwOut{$\widehat{\Delta}(A, A')$}
\Indp%
    $\bar{g}_A \gets \frac{1}{|A|} \sum_{j \in A} g_j$ \tcp*{{maintained incrementally}}
    $\bar{g}_{A'} \gets \frac{|A| \cdot \bar{g}_A + g_i}{|A| + 1}$ \tcp*{one vector addition}
    \uIf{task is fairness-aware}{
        $s(A) \gets \big(1 - \tfrac{\lambda}{100}\big) \cos\big(g_\target^{\mathrm{acc}},\, \bar{g}_A^{\mathrm{acc}}\big) + \tfrac{\lambda}{100} \cos\big(g_\target^{\mathrm{fair}},\, \bar{g}_A^{\mathrm{fair}}\big)$\;
        $s(A') \gets \big(1 - \tfrac{\lambda}{100}\big) \cos\big(g_\target^{\mathrm{acc}},\, \bar{g}_{A'}^{\mathrm{acc}}\big) + \tfrac{\lambda}{100} \cos\big(g_\target^{\mathrm{fair}},\, \bar{g}_{A'}^{\mathrm{fair}}\big)$\;
    }
    \Else{
        $s(A) \gets \cos\big(g_\target,\, \bar{g}_A\big)$\;
        $s(A') \gets \cos\big(g_\target,\, \bar{g}_{A'}\big)$\;
    }
    $\widehat{\Delta}(A, A') \gets s(A') - s(A)$\;
    \Return $\widehat{\Delta}(A, A')$\;
\Indm%
\end{algorithm}

\noindent\textbf{Scoring Subsets by Gradient Alignment.}
We score a subset $A$ by the cosine similarity between mean of its source gradient and  target gradient: $\mathrm{score}(A) = \cos(g_\mathcal{T},\frac{1}{|A|}\sum_{i \in A} g_i).$ 
Marginal gain of adding source $S_i$ to $A$ is  estimated as 
$  \widehat{\Delta}(A, S_i) = \mathrm{score}(A \cup \{S_i\}) - \mathrm{score}(A).
$
Because gradients are additive, $\mathrm{score}(A \cup \{S_i\})$ is computed incrementally by updating the  mean gradient, so each candidate evaluation after the first costs only two dot products.
For fairness-aware tasks, we keep  accuracy and TPR-parity gradients separate and combine their cosine scores. Combining gradients $g^{\mathrm{acc}}$ and $g^{\mathrm{fair}}$ can lead to misleading directions when objectives are correlated or have heterogeneous scales~\cite{sener2018multi} while combining cosines preserves each objective's direction-alignment signal independently. Fairness weight $\lambda$ tunes the trade-off between accuracy and fairness.

\vspace{1mm}\noindent\textbf{Complexity.}
Precomputation requires one warm-start training pass 
and one gradient evaluation per source and target 
i.e.\ $O(n)$ gradient evaluations 
while online queries cost $O(d_g)$ per candidate move ($d_g$ ranging from a few tens of parameters for tabular tasks to roughly $d_\mathrm{feat}\cdot c$ for image classification with $c$ categories).
Table~\ref{tab:oracle_cost} summarizes  cost trade-offs of the marginal utility oracles.

\begin{table}[h]
  \centering
  \caption{Cost comparison of marginal gain oracles. $T_{\text{ref}}$: cost of fitting a reference model (one linear/small head) ($<< T_{\text{train}}$).}
  \label{tab:oracle_cost}
  \small
  \setlength{\tabcolsep}{5pt}
  \begin{tabular}{lll}
    \toprule
    Oracle & Offline cost & Per-query \\
    \midrule
    Exact (\sys) &
      None & $O(T_{\text{train}})$ \\
    Data-profile (\sysdp) &
      $O(n \cdot T_{\text{train}})$ & $O(d_\phi)$ \\
    Gradient-match (\sysgrad) &
      $O((n\!+\!1) \cdot T_{\text{ref}})$ & $O(d_g)$ \\
    \bottomrule
  \end{tabular}
\end{table}

\section{Experimental Evaluation}
\label{sec:exp}

We evaluate \sys and its variants against  baselines over seven benchmarks spanning structured fairness-aware classification, regression, image classification under location shift and text sentiment analysis. We  address the following questions: (\textbf{RQ1}) How does source selection compare to baselines for improving utility over diverse tasks? (\textbf{RQ2}) How does the efficiency of \sys compare to baselines? (\textbf{RQ3}) Do the variants of \sys preserve solution quality while reducing evaluation costs? (\textbf{RQ4}) How do \sys and its variants scale as the number of sources grows?
\begin{table*}[t]
\small
\caption{Profit by task and dataset, mean $\pm$ 95\% CI over 30 target splits
(ACS) or 10 model seeds (Wilds, Amazon). ACSTravelTime reports MSE (lower is
better); all other columns are higher-is-better. Best per column in {\color{gray}bold};
Second best are \underline{underlined}.}
\begin{tabular}{lccccc}
\toprule
\textbf{Method} & \textbf{ACSIncome} & \textbf{ACSPubCov.} & \textbf{ACSTravelTime} & \textbf{Amazon}& \textbf{Wilds} \\
\midrule
\textbf{\textsc{Single}}      & 72.81 $\pm$ 0.22 & 62.55 $\pm$ 0.04 & 500.67 $\pm$ 1.33 & 0.8337 $\pm$ 0.0071 & 0.0903 $\pm$ 0.0077 \\
\textbf{\textsc{AllSrc} }     & 71.39 $\pm$ 0.21 & 51.12 $\pm$ 0.03 & 500.28 $\pm$ 1.31 & 0.8314 $\pm$ 0.0118 & 0.2114 $\pm$ 0.0096 \\
\textbf{\textsc{Random}}      & 70.17 $\pm$ 0.18 & 51.68 $\pm$ 0.02 & 609.28 $\pm$ 1.49 & 0.8163 $\pm$ 0.0108 & 0.1408 $\pm$ 0.0119 \\
\textbf{\textsc{Greedy}}     & 72.98 $\pm$ 0.28 & 62.55 $\pm$ 0.04 & 496.42 $\pm$ 1.28 & 0.8470 $\pm$ 0.0066 & 0.1463 $\pm$ 0.0056 \\
\textbf{\textsc{DsDM}}        & 72.81 $\pm$ 0.19 & 64.61 $\pm$ 0.04 & 496.72 $\pm$ 1.27 & 0.8550 $\pm$ 0.0069 & 0.2132 $\pm$ 0.0072 \\
\textbf{\textsc{Grasp}}       & 73.13 $\pm$ 0.21 & 64.76 $\pm$ 0.04 & \textbf{484.45 $\pm$ 1.29} & \textbf{0.8614 $\pm$ 0.0054} & \underline{0.2303} \\
\midrule
\textbf{\textsc{Splice}}      & \underline{73.14 $\pm$ 0.22} & \underline{64.81 $\pm$ 0.04} & \underline{485.44 $\pm$ 1.28} & \underline{0.8613 $\pm$ 0.0055} & \textbf{0.2329 $\pm$ 0.0074} \\
\textbf{\textsc{Splice-DP}}   & \textbf{73.20 $\pm$ 0.22} & 64.73 $\pm$ 0.05 & 485.87 $\pm$ 1.27 & 0.8549 $\pm$ 0.0059 & 0.2219 $\pm$ 0.0068 \\
\textbf{\textsc{Splice-Grad}} & 72.93 $\pm$ 0.28 & \textbf{64.87 $\pm$ 0.04} & 486.71 $\pm$ 1.24 & 0.8518 $\pm$ 0.0054 & 0.2241 $\pm$ 0.0076 \\
\bottomrule
\end{tabular}
\label{tab:results-by-task}
\end{table*}
\subsection{Experimental Setup}
\label{sec:setup}

\noindent\textbf{Benchmarks.}
We use seven benchmarks across four families:
\vspace{1mm}\noindent(a)~\underline{\textsf{SANTOS UK government spending tables~\cite{khatiwada2023santos}.}} We consider two real-world benchmarks
derived directly from \santos discovery~\cite{khatiwada2023santos}.
\textbf{YDNPA} and \textbf{IPO} have 20 monthly spending files from Yorkshire Dales National Park Authority and UK Intellectual Property Office, respectively. Tables in a group share an identical schema;  
we use one table per group as target
and 19 as candidate pool, consider binary classification task of predicting if a transaction exceeds median spend
and  
report binary F1 score on held-out target data.
(b)~\underline{\textsf{ACS Folktables~\cite{ding2021retiring}.}} We consider three datasets partitioned by U.S. states into $51$ sources: \textbf{ACSIncome} for binary classification to predict if an individual earns above $\$50$k; with sex as  protected attribute, we use profit function $\mathcal{P}_\mathcal{T} = Accuracy + \lambda \cdot TPR$, $\lambda = 50$. \textbf{ACSPublicCoverage}  for binary classification to predict if an individual has public health insurance coverage ($\lambda = 0$). \textbf{ACSTravelTime} for regression  to predict commute time
and maximize $\profit_\target(S) = -\mathrm{MSE}(h_S, \target)$.
We evaluate our solutions with ground truth on a diverse selection of $15$ 
states as candidate sources and Puerto Rico as held-out target (30/70 validation-test split); 
we report our results on the entire dataset in Section~\ref{sec:scalability}.
(c)~\underline{\textsf{Wilds~\cite{koh2021wilds}}}
contains wildlife images 
obtained from $30$ camera-trap locations as candidate sources (
each with $>2$k images). We randomly hold out $3$ additional sources as target (30/70 validation-test split)
and consider multi-class image classification ($c = 80$ species, image features extracted with a frozen ImageNet-pretrained ResNet-18 with 512-dim embeddings). 
We report macro-F1 score on test  to account for long-tail class imbalance in camera-trap data.
(d)~\underline{\textsf{Amazon Reviews~\cite{hou2024bridging}}}  contains reviews of products cataloged on Amazon; 
we sample $270$k reviews: 
24 categories
as candidates and 3 
as target (30/70 validation-test). We consider a binary sentiment classification task: a review is positive if its rating $\ge 4$, and report binary F1. 

\vspace{1mm}\noindent\underline{\textsf{SynPubCov (scalability).}} To perform scalability experiments in Section \ref{sec:scalability}, we build a synthetic benchmark from \textsf{ACSPublicCoverage} by sampling $1,000$ sources each containing $5,000$ examples via stratified resampling  
with controlled label distribution skew.

\vspace{1mm}\noindent\noindent\textbf{Downstream models.}  
We train a logistic regression model for classification tasks, a linear regression model for regression tasks, a linear head for image classification, and a small MLP on frozen features for sentiment analysis. This choice of fast models is deliberate as it isolates source-selection quality from model capacity. For the \santos benchmarks, we train a logistic regression model with $C{=}0.1$ and $500$ max iterations. {\sysgrad uses a separate, lightweight reference model for gradient extraction (Appendix B), which is distinct from the downstream model used for utility evaluation.}

\vspace{1mm}\noindent\noindent\textbf{Competing methods.}
\label{sec:baselines}
We consider following methods:
(1)~\underline{\textsc{Single}} evaluates each source independently and selects one with highest individual profit.
(2)~\underline{\textsc{AllSrc}}: na\"ive baseline that pools all sources into a single training set.
(3)~\underline{\textsc{Random}} trains on a randomly sampled subset, averaged over $10$ independent runs.
(4)~\underline{\textsc{Greedy}} iteratively adds sources with  highest individual profit score, stopping when no further improvement is found. 
(5)~\underline{\textsc{Grasp}}~\cite{bermejo2011grasp}: randomized greedy construction plus local-search refinement, with construction restricted to a top-$k$ candidate list.
(6)~\underline{\textsc{DsDM}}~\cite{engstrom2024dsdm}: source-level adaptation of datamodels; we train models on subsets corresponding to each of $N$ binary subset indicator vectors, fit a regression model to learn profits, rank and greedily select sources by coefficient.
\noindent\underline{\sys}, \underline{\sysdp}, \underline{\sysgrad}: our solutions
(Sections~\ref{sec:approach} and~\ref{sec:estimation}).

\vspace{1mm}\noindent\textbf{Parameter settings.} We set the following parameters.
ACS $B=7$, Amazon $B=10$, Wilds $B=15$; $k_{\max}=B$ on benchmarks in Table~\ref{tab:results-by-task}. \grasp  $r=20$ randomized constructions, restricted candidate list capped at $k=5$. \textsc{DsDM}, we sample $N=500$
distinct non-empty binary subset indicator vectors and train a model per
subset to create  surrogate dataset. \random averages $10$ independent draws with sizes uniform on $\{1,\dots,B\}$.
\sysdp meta-dataset constructed with $m=5$ additions and $m=5$ removals per source; 
standardized linear regression on profile differences fitted for tabular data and gradient-boosting regressor ($300$ trees, depth $5$, learning rate $0.05$) fitted on  richer meta-feature vector
for image and text data.
\sysgrad warm-starts  reference model for $5$ epochs on  union of all sources before extracting gradients (per-task architectures in Appendix B). Fairness weight $\lambda=50$ on ACSIncome; $0$ elsewhere.
We use seed $12345$ for ACS, $42$ for others. Model trainings, baselines, and \sys variants share the same evaluation model 
so that profit differences reflect selection quality alone. Reported runtimes are sequential unless explicitly stated as  parallel 
(as in Section~\ref{sec:scalability}).
All experiments run on Apple MacBook M3 Pro with 36 GB RAM.

\vspace{1mm}\noindent\textbf{Performance metrics.}
We report three complementary measures: \textit{profit} on the held-out target, \textit{wall-clock time} in seconds, and the number of \textit{models explored} or trained. 

\subsection{Effectiveness of \sys}
\label{sec: effectiveness}

We report profit scores of the competing methods across five datasets in Table~\ref{tab:results-by-task} addressing \textbf{RQ1} and \textbf{RQ3}.
We report mean with $95\%$ confidence intervals over $30$ independent target splits on ACS benchmarks and $10$ model seeds on Wilds and Amazon. We further perform paired testing to compare close pairs on identical splits (Table 10, Appendix C) because split-to-split variance is large and shared across methods and marginal intervals overlap even where a method wins on every split. On four of the five benchmarks, a \sys variant is either the best or statistically indistinguishable from the best (exception being ACSTravelTime with \grasp  
leading by a $0.2\%$ with $12.5\times$ more model evaluations). Several findings stand out:

\noindent\textit{\underline{Consolidating all sources consistently underperforms}.} 
On ACSPublicCoverage, 
\allsrc
yields a profit of 51.12, a striking 13.69 points below \sys's 64.81 and below even random selection. This behavior is due to source cancellation in its extreme form: a handful of sources with low accuracies dominate the pooled model suppressing overall accuracy. On ACSTravelTime, \textsc{AllSrc} has higher MSE (500.28) compared to \sys (485.44); on Amazon, this gap is $\approx3$ F1 points (0.8314 vs 0.8613).
On Wilds, surprisingly, \allsrc (0.2114) substantially outperforms \single (0.0903), which is consistent with the structure of the task: each individual camera trap location captures only one habitat/narrow species set and pooling $30$ locations provides breadth. 
Selection still helps -- \sys adds a further 2.2 macro-F1 points to \allsrc -- 
no individual source is harmful enough to drag the model toward a degenerate prediction. The lesson: ``more data is not necessarily helpful'' is task-dependent; \sys dominates whether or not \allsrc is competitive.

\noindent\textit{\underline{\sys variants preserve quality}}. \sysdp outperforms other me- thods on ACSIncome ($73.20$) while \sysgrad leads on ACSPublicCoverage ($64.87$: never behind \sys on any of the $30$ splits, and strictly ahead on $21$, \textit{p}$<10^{-4}$). On Wilds, where exact evaluation is most expensive, both variants are statistically tied with \sys(\textit{p}=$0.18$) while using $16-20\times$ fewer evaluations. The deficit is small where the variants do trail: at most $1.26$ MSE on ACSTravelTime ($0.26\%$) and $0.011$ macro-F1 on Wilds ($4.7\%$, not significant). The variants combine \sys's advantage over \grasp to reach its competitive quality at orders of magnitude fewer model trainings.

\noindent\textit{\underline{\grasp is effective but costly}.} \sys and \grasp are statistically indistinguishable on ACSIncome (\textit{p}$=0.84$, returning identical subsets on 20/30 splits) and Amazon (\textit{p}$=0.47$); \sys leads on ACSPublicCoverage by $0.044$ (\textit{p}$=0.005$) while \grasp leads on ACSTravelTime by $0.99$MSE $(p<10^{-4})$. With $319-6,385$ model evaluations, \sys achieves $12.5-23.5\times$ reduction against \grasp's $5,196-30,974$ trainings. On Wilds \grasp needs 14,606 trainings in $12.4$ hours, reinstating its infeasibility for tasks where exact retraining is crucial.

\vspace{1mm}\noindent\textbf{Comparing selected sources with discarded sources.}
\label{sec:characterize}
To better understand  characteristics of the  sources selected by \sys that distinguish them from  discarded sources, we profile each candidate source in terms of their size, label mismatch against target, distributional shift, feature coverage, and profit with single source training (details in Appendix D).
We observe that no individual source characteristic consistently indicates whether a source is selected.
Label-distribution mismatch is starkest on IPO (worst source has $32.4\%$ of transactions above target median against target's $50\%$ and attains $F_1=0$ as in Section \ref{sec:santos}). Feature coverage gaps dominate Wilds where $7$ locations contain $2-6$ classes covering at most $13$\% of  target classes. Source cancellation drives ACSPubCov where pooling all sources scores $51.12$ against \sys's $64.81$.
In ACSPubCov, California data has the highest solo profit ($62.2$) but a marginal contribution of $-5.89$ and \sys discards it; New York data ranks ninth by solo profit ($54.2$) and contributes $+7.77$. 
This behavior is evident in Wilds where sources selected by \sys have \emph{lower} individual profit than those it discards ($0.029$ vs $0.046$), cover fewer target classes ($0.258$ vs $0.423$), and are \emph{further} from the target ($9.28$ vs $6.81$). \sys selects six of seven locations that have a common macro-F1 of $0.0184$ and contain $2$--$6$ classes. These sources 
represent a region of label space sampled thinly by the broad-coverage locations: 
deciding which sources to select depends on sources that are already selected.
The pattern is also visible in Amazon where \sys selects $3$ of $24$ categories and top-2 sources with  highest individual profits  
($0.829$ and  
$0.817$) have low label mismatch (0.027 and 0.003) and are discarded.

\vspace{1mm}\noindent\textbf{Generalizability to downstream learner.}
\label{sec:learners}
\sys scores candidate subsets with  downstream model, so a natural
question is whether its choices survive a change of learner. We address this question in two
modes: (a)~ \emph{transfer} mode selects sources using linear oracle of Table~\ref{tab:results-by-task} and trains returned
subset on XGBoost and LightGBM;
(b)~\emph{native} mode re-runs search with boosted learner
as its oracle. Table~\ref{tab:learners} reports both modes.
\begin{table}[h]
  \centering \scriptsize
  \setlength{\tabcolsep}{2pt}
  \caption{Generalizability to downstream learner.  
  Bold=best.}
  \label{tab:learners}
  \begin{tabular}{llcccccc}
    \toprule
    Dataset & Learner & \allsrc & \single & \greedy & \textsc{DsDM} & \grasp & \sys \\
    \midrule
    \multirow{5}{*}{ACSInc.}
      & LogReg         & 71.39 & 72.81 & 72.98 & 72.81 & 73.13 & \textbf{73.14} \\
      & XGBoost (transfer)       & 78.71 & \textbf{82.21} & 81.18 & 77.54 & 80.02 & 80.16 \\
      & LightGBM (transfer)       & 78.63 & \textbf{82.39} & 81.17 & 77.94 & 79.93 & 80.03 \\
      & XGBoost (nat)  & 79.33 & 82.73 & \textbf{83.59} & 82.73 & --- & 81.41 \\
      & LightGBM (nat) & 79.12 & \textbf{83.45} & 82.69 & 83.11 & --- & 82.25 \\
    \midrule
    \multirow{5}{*}{ACSPub.}
      & LogReg         & 51.12 & 62.55 & 62.55 & 64.61 & 64.76 & \textbf{64.81} \\
      & XGBoost (transfer)       & 64.20 & 63.52 & 63.52 & 67.76 & 67.77 & \textbf{68.62} \\
      & LightGBM (transfer)      & 65.58 & 64.28 & 64.28 & 68.09 & 68.11 & \textbf{69.04} \\
      & XGBoost (nat)  & 64.19 & 69.31 & 69.31 & 67.90 & --- & \textbf{69.74} \\
      & LightGBM (nat) & 65.73 & 69.09 & 69.09 & 68.53 & --- & \textbf{70.18} \\
    \midrule
    \multirow{5}{*}{ACSTrvl.}
      & LinReg         & 500.28 & 500.67 & 496.42 & 496.72 & \textbf{484.45} & 485.44 \\
      & XGBoost (transfer)       & 440.49 & 478.16 & 468.86 & 468.57 & 433.05 & \textbf{423.47} \\
      & LightGBM (transfer)      & 430.21 & 493.55 & 429.90 & 458.08 & 431.06 & \textbf{428.72} \\
      & XGBoost (nat)  & 439.07 & 437.31 & 419.52 & 412.50 & --- & \textbf{408.02} \\
      & LightGBM (nat) & 436.17 & 436.43 & 422.71 & 417.00 & --- & \textbf{400.74} \\
    \bottomrule
  \end{tabular}
\end{table}
We find that on some datasets (
ACSPubCov, ACSTravelTime) the linear-selected subset remains the best choice
under both boosted learners, and by wider margins than under the oracle it was
selected with: \sys{} beats \allsrc{} by $4.41$ points on ACSPubCov and by
$17.0$ MSE on ACSTravelTime under XGBoost. 
Re-running the
search with the boosted learner as its own oracle reproduces the ordering on a
single target split: \sys{} leads on both the datasets under
both learners. Each initialization costs mins rather than seconds, so the
cheap oracle remains the better operating point, and we report native results for base split.

We observe a different behavior on ACSIncome where the ranking reverses under both boosted learners. Training on the single best source
beats \sys by around two points. Note that both subsets were chosen under the linear oracle, so \sys's guarantee of dominating the best single source applies to the model it searched with, not to the one it is retrained under. Decomposing profit into its two terms explains the reversal. This best source is \emph{always} the more accurate choice and \emph{always} the less fair one, so under the linear model its accuracy advantage is outweighed by a parity penalty scaled by $\lambda=50$. Boosted learners close most of that parity gap on their own
while the accuracy advantage remains, and the ranking flips. The subset returned by \sys 
was selected under a different objective than what it is scored on. 
Re-running the search natively
optimizes the same mis-specified objective more effectively, so the binding constraint is the profit function rather than the oracle.

\subsection{Efficiency of \sys }
\label{sec:efficient}

\begin{figure}[t]{}
    \centering
    \includegraphics[width = \linewidth]{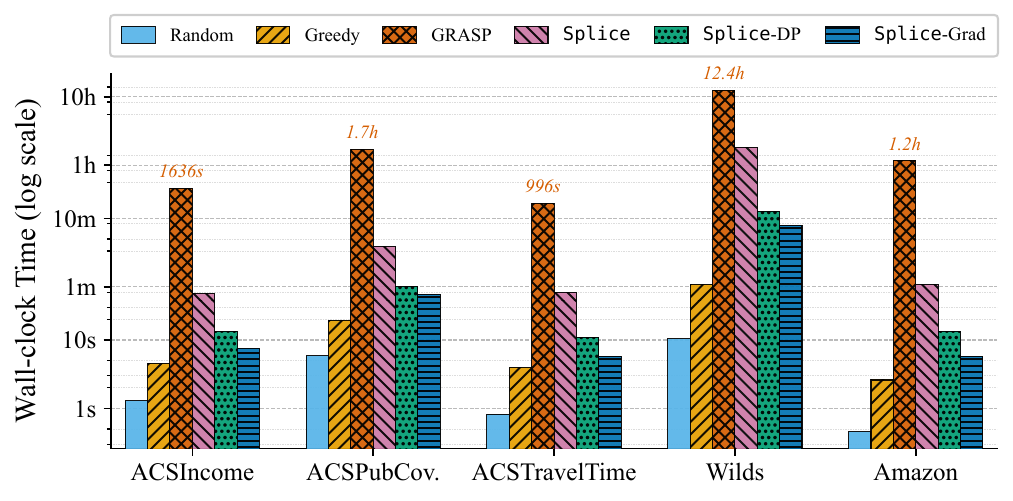}
    \captionof{figure}{Runtime of methods (in seconds) on different tasks. 
    }
    \label{fig:method_time}
\end{figure}

\begin{figure}[t]{}
    \centering
    \includegraphics[width = \linewidth]{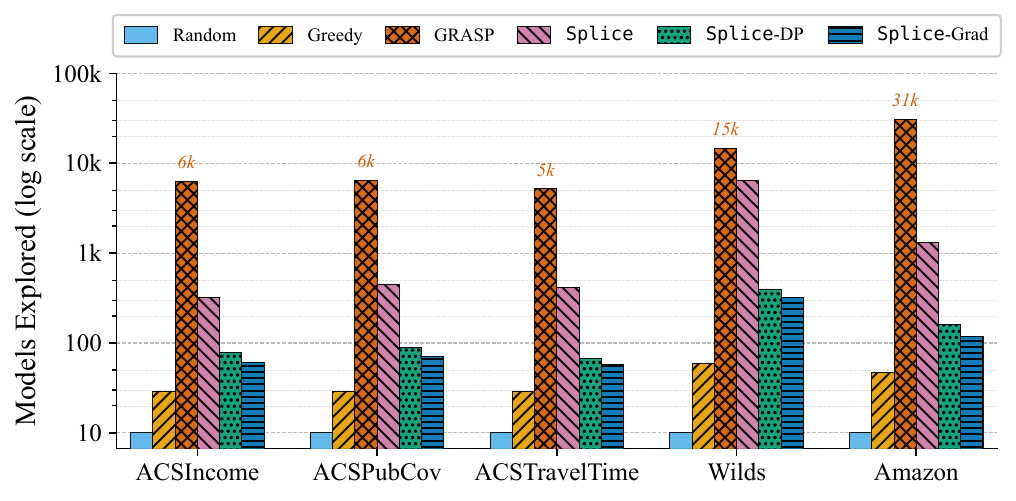}
    \captionof{figure}{Number of models explored on different tasks.}
    \label{fig:method_modex}
\end{figure}

Figures~\ref{fig:method_time} and~\ref{fig:method_modex} address \textbf{RQ2}, comparing wall-clock time and model evaluations across all methods and five datasets. We observe that \grasp is prohibitively expensive,
requiring $6,259$ model evaluations, $27$ mins of wall-clock time on ACSIncome, $31k$ evaluations, $4,185$s on Amazon and $14.6k$ evaluations, $>12$ hrs on Wilds. \sys meets/outperforms with $319- 6,385$ evaluations in $47$s-$1.8$ hrs
(upto $23.5\times$ reduction in evaluations and $6.7-64\times$ in wall-clock time). 

\noindent\textit{\underline{\sys variants achieve near-\sys quality} but at \greedy-level cost} 
requiring 57-394 model evaluations across all datasets, comparable to \greedy ($29-59$ evaluations), but with substantially higher final profit. On Wilds, \sysgrad ($322$ evaluations, macro-F1 $0.224$) outperforms \greedy ($59$ evaluations, $0.1463$) for a $5\times$ increase in evaluations -- a favorable trade-off where each evaluation involves training a neural head. Runtime plot Figure~\ref{fig:method_time} confirms the observation that 
\sysdp and \sysgrad occupy the same runtime tier as \greedy while delivering profit scores closer to \sys.

\begin{figure}[t]
    \centering
\includegraphics[width=\linewidth]{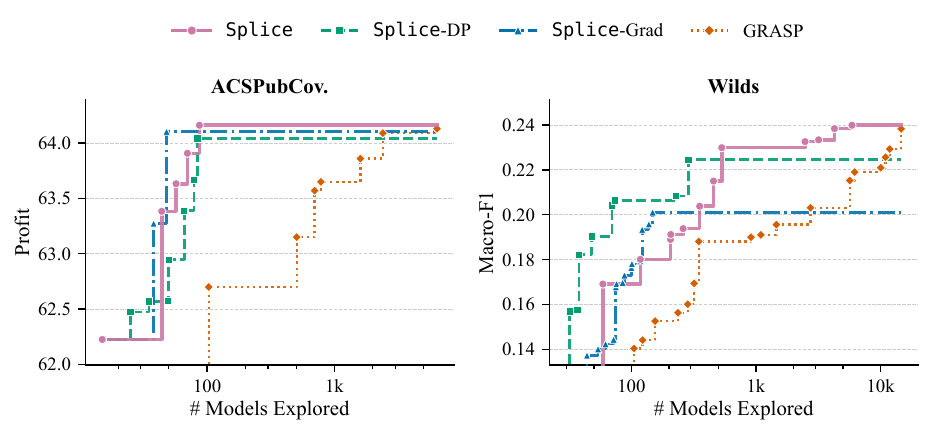}
    \caption{Effect of number of models explored on profit.}
    \label{fig:method_history}
\end{figure}

Figure~\ref{fig:method_history} shows `anytime' performance, the best profit found as a function of models explored, addressing how quickly each method finds a good solution. Convergence is consistent across tasks: \sysdp and \sysgrad jump sharply to near optimal profit within $50-300$ evaluations before plateauing, trading a small amount of final quality for fast convergence. \sys starts lower because each step pays the cost of an exact training, overtakes the variants at around $500$ evaluations on Wilds, and continues to improve while the methods converge on ACSPublicCoverage. \grasp makes slow, stepwise progress through thousands of evaluations before approaching \sys's performance.

On ACSPublicCoverage, \sysgrad converges in $<50$ evaluations, reflecting gradient alignment as a reliable signal when target feature distribution is well-separated between classes. \sys continues improving past $1000$ evaluations on Wilds, eventually reaching macro-F1 = $0.240$, because the source pool contains complementary subsets that become apparent only when larger initialization sizes are explored. \sys variants plateau earlier, 
reflecting that gradient and profile signals identify the most obvious compatible locations quickly but miss useful combinations that deeper search uncovers.

\noindent{\textit{\underline{Task-structure dependence of surrogate strengths}.} The relative effectiveness of the variants varies with task structure. \sysdp performs best on tabular tasks where distributional statistics are directly informative and profile differences correlate 
with marginal gain — across 
ACS tasks, \sysdp matches \sys within 0.43 MSE points on regression and within 0.08 
points on classification. \sysgrad performs best on tasks where a natural neural reference model exists and gradient directions encode task-specific signal (e.g., metric-aware soft losses for macro-F1 or binary-F1 objectives). On Wilds, \sysgrad outperforms \sysdp by 0.2 macro-F1 pts while the two are competitive  on text classification. 
}

\noindent\textit{\underline{Practical recommendation}.} We conclude that \sys variants are effective when evaluation budgets are tight or  training is expensive whereas \sys is preferred when a few hundred model evaluations are affordable
and quality is critical. {\sysgrad is effective in image classification with frozen-backbone features and tasks where a neural reference model is natural. \sysdp is recommended for tabular tasks or when no suitable reference model exists}.

\subsection{Scalability analysis}
\label{sec:scalability}

We study how well \sys scales (\textbf{RQ4}) along two dimensions: (a)~reporting the results on the entire ACS dataset comprising $50$ states; and (b)~increasing the number of sources from $10$ to $1,000$.

\subsubsection{Scaling to ACS data ($50$ states)}
Due to the prohibitive computation of baselines, we evaluate \sys and its variants on all $50$ states of ACS data as candidate sources
(Table~\ref{tab:acs50}). 
\begin{table}[h]
  \centering
  \caption{Results on the entire ACS data ($50$ sources)
    \#Evals = full model trainings; parallel speedup in parentheses.
    }
  \label{tab:acs50}
  \small
  \setlength{\tabcolsep}{1.1pt}
  \begin{tabular}{llcccl}
    \toprule
    Dataset & Method & Profit & \#Evals
            & Sequential (s) & Parallel (s) \\
    \midrule
    \multirow{3}{*}{ACSIncome}
      & \sys     & 74.610 &  8{,}353 & 4{,}979s & 1{,}155s (4.3$\times$) \\
      & \sysdp   & \textbf{74.778} &     367  &   522s   &     92s (5.7$\times$) \\
      & \sysgrad & 74.318 &     326  &   243s   &     51s (4.8$\times$) \\
    \midrule
    \multirow{3}{*}{ACSPubCov}
      & \sys     & \textbf{66.825} & 11{,}030 & 6{,}274s & 1{,}216s (5.2$\times$) \\
      & \sysdp   & 65.845 &     401  &   552s   &    158s (3.5$\times$) \\
      & \sysgrad & 65.677 &     323  &   375s   &     71s (5.2$\times$) \\
    \midrule
    \multirow{3}{*}{ACSTrvlTime}
      & \sys     & \textbf{477.156} & 16{,}808 & 3{,}982s & 627s (6.3$\times$) \\
      & \sysdp   & \textbf{477.156} &     244  &   209s   &  44s (4.8$\times$) \\
      & \sysgrad & \textbf{477.156} &     219  &    51s   &  14s (3.6$\times$) \\
    \bottomrule
  \end{tabular}
\end{table}

\noindent\textit{\underline{More sources improve profit}.}
Expanding the pool to $50$ states yields consistent profit improvements: gains are most pronounced on ACSPublicCoverage: \sys improves by $+2.0$ points 
and the surrogates by $+0.8$ to $+1.1$, confirming that additional $35$ sources contain useful signals beyond the $15$ selected states. ACSTravelTime improves modestly ($8.3-9.6$ points), consistent with the task having already captured most of the signal at 15 sources. Expanding the pool, therefore, augments \sys performance. 

\begin{figure}[]{}
    \centering
    \includegraphics[width = \linewidth]{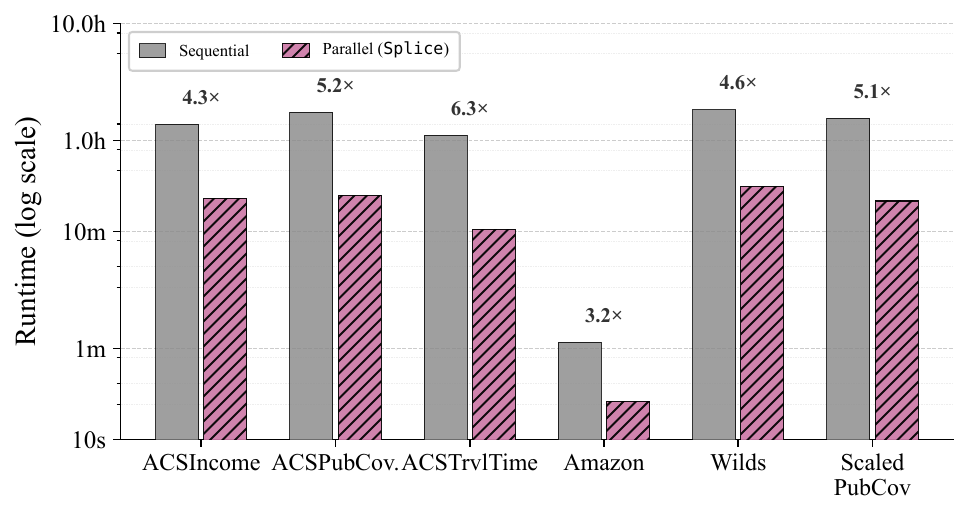}
    \captionof{figure}{\sys runtime speedup due to parallelization.
    }
    \label{fig:scaling_parallel}
\end{figure}
 
\noindent\textit{\underline{Surrogate evaluation reduction grows with pool size}.}
Model evaluations further reduced to 22.8-76.7$\times$ at 50 sources. 
The number of candidate swap moves grows quadratically with pool size making exact evaluation increasingly expensive (\sys requires $8.4k-16.8k$ evaluations) while surrogate query cost remains mostly constant (\sysdp needs $244-401$ and \sysgrad $219-326$ evaluations).
 
\noindent\textit{\underline{Parallelization is uniformly effective}.}
Figure~\ref{fig:scaling_parallel} shows that \sys has 4.3-6.3$\times$ parallel speedup
at 50 sources, reducing wall-clock time from 1.1-1.75 hrs to 10-20 mins. \sysdp achieves 3.5-5.7$\times$ speedup and \sysgrad 3.6-5.2$\times$. Across all six benchmarks the speedup holds within a factor of two (3.2–6.3$\times$) despite evaluation cost varying by three orders of magnitude.
Exact selection over large number of sources is thus an interactive-time operation.

\subsubsection{Synthetic scaling: surrogates are essential past $\sim$100 sources} 
\begin{figure}[t]
     \centering
         \begin{minipage}{\linewidth}
        \centering
        \includegraphics[width=0.6\textwidth]{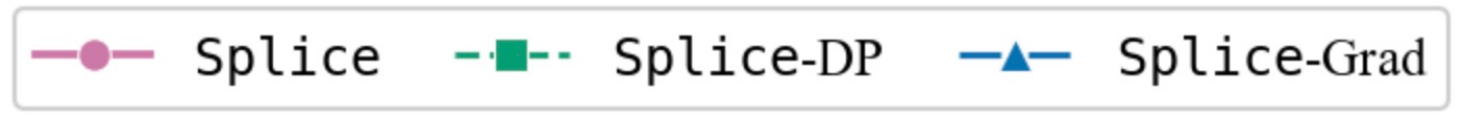} 
    \end{minipage}
     \begin{subfigure}{0.32\linewidth}
         \centering
         \includegraphics[width=\linewidth]{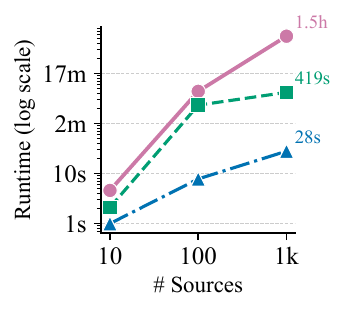}
         \caption{Runtime.}
         \label{fig:synthetic_runtime}
     \end{subfigure}
     \begin{subfigure}{0.32\linewidth}
         \centering
        \includegraphics[width=\linewidth]{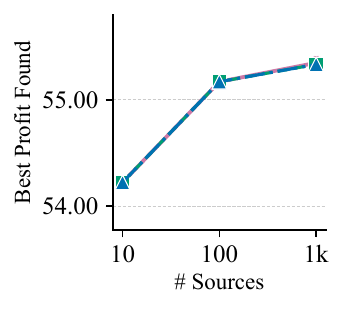}
         \caption{Solution quality.}
         \label{fig:synthetic_profit}
     \end{subfigure}
     \centering
     \begin{subfigure}{0.32\linewidth}
         \centering
         \includegraphics[width=\linewidth]{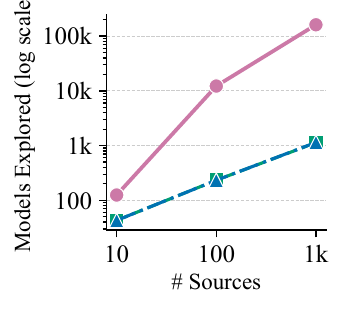}
         \caption{Models explored.}
         \label{fig:synthetic_modex}
     \end{subfigure}
        \caption{Effect of number of sources
        (SynPubCov).
        }
        \label{fig:synthetic}
\end{figure}

\sys runtime grows from 4.6s at n=10 sources to 1.6 hrs at n=1,000 (Figure~\ref{fig:synthetic_runtime}) because each swap evaluates a number of candidates proportional to pool size. \sysdp takes 7 mins and \sysgrad 28s -- a $13\times$ and 199$\times$ reduction respectively in runtime. \sysgrad is performant with increasing pool size since its per query cost is dominated by a vector operation rather than model training. Despite the cost gap, \sysgrad returns the same subset as \sys at n=10 and n=100 using $2.9\times$ and $52.7\times$ fewer evaluations (Figure~\ref{fig:synthetic_profit}), confirming that it does not sacrifice quality for efficiency.
\begin{figure*}[t]
     \centering
    \begin{minipage}{\linewidth}
        \centering
        \includegraphics[width=0.3\textwidth]{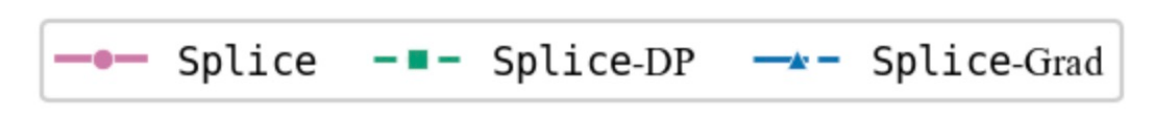} 
    \end{minipage}
     \begin{subfigure}{0.49\linewidth}
         \centering        \includegraphics[width=\linewidth]{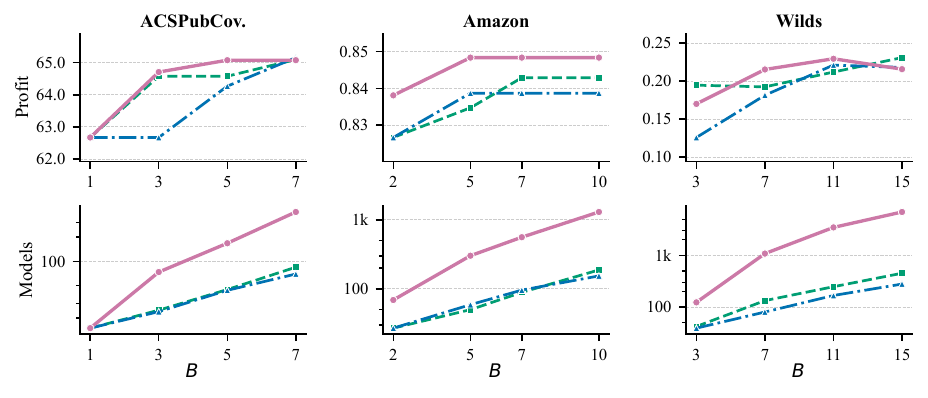}
         \caption{Varying maximum subset size, B.}
         \label{fig:ablation_smax}
     \end{subfigure}
     \centering
     \begin{subfigure}{0.49\linewidth}
         \centering
        \includegraphics[width=\linewidth]{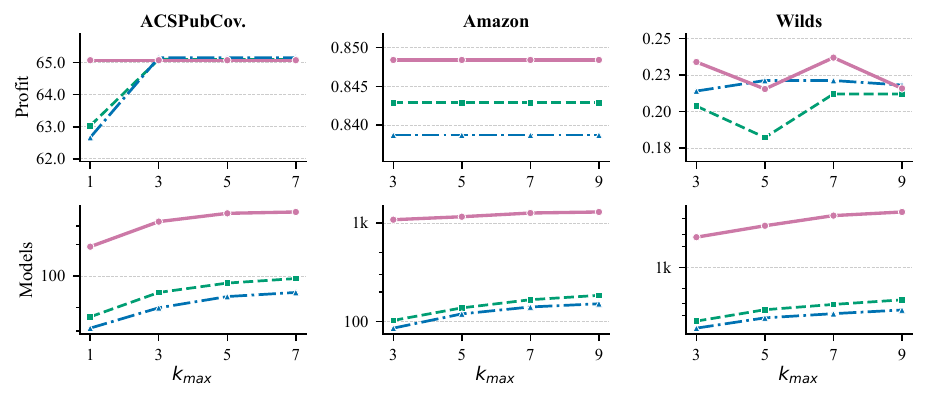}
         \caption{Varying size of swaps, kmax.}
         \label{fig:ablation_kmax}
     \end{subfigure}
        \caption{Sensitivity of \sys to budget B and maximum number of swaps $k_{max}$.
        }
        \label{fig:ablation}
\end{figure*}
At $n=1,000$ sources (Figure~\ref{fig:synthetic_modex}), \sys explores $>161$k subsets  compared to 
$1,140$ explored by its variants. This two-orders-of-magnitude reduction in evaluation count directly translates to the observed runtime savings. For large source pools, \sys variants are not just faster, but the only practical option.
\subsection{ 
Integration with Data Discovery}
\label{sec:santos}
Data discovery tools narrow an open-ended corpus to a tractable set of schema-unionable tables, but do not reason about their utility for the downstream task. We quantify that gap on two real-world government spending benchmarks (Table~\ref{tab:santos}).

\begin{table}[h]
  \centering
  \caption{\sys on table groups discovered by \santos. `Self-train UB': F1 attained by 5-fold CV on target table. 
  }
  \label{tab:santos}
    \resizebox{\linewidth}{!}{
  \begin{tabular}{lccc}
    \toprule
    Dataset & SANTOS + Single & SANTOS + All & SANTOS + \sys \\
    \midrule
    YDNPA (Spending) & 0.737 & 0.722 & \textbf{0.768} \\
    IPO (Payments)   & 0.736 & 0.715 & \textbf{0.745} \\
    \bottomrule
  \end{tabular}}
  \begin{tablenotes}
    \small
    \item Self-train UB (YDNPA): 0.737 \ Self-train UB (IPO): 0.726
  \end{tablenotes}
\end{table}

\noindent\textbf{The all-sources baseline fails.} Training a logistic regression classifier on all $19$ sources output by \santos yields F1=0.722 on YDNPA and F1=0.715 on IPO. These results fall below the self-training upper bound (0.737 and 0.726, respectively), implying that using all the tables is \emph{worse} than training on the target table alone. Source-level evaluation reveals that individual source qualities vary substantially, ranging from 0.584 to 0.737 on YDNPA. The spread is more dramatic on IPO with the worst source achieving F1 = 0.000, because its spending distribution ($32.4\%$ above target median) diverges sharply from the target ($50\%$ above target median), causing the classifier to predict nearly every transaction as low spend. While \santos correctly identifies all 19 tables as schema-related, it cannot identify which of the tables are relevant to the task.

\noindent\textbf{\sys recovers and exceeds the upper bound.} Applying \sys to each source pool selects a compact subset that significantly outperforms the all-sources baseline. On YDNPA, a 6-table subset 
achieves F1 = 0.768 (a gain of 4.6 points over all sources and 3.1 points over self-training upper bound) while on IPO, a 7-table subset 
achieves F1 = 0.745 (a gain of 3.0 points over all sources and 1.9 points over upper bound). Exceeding the self-training bound is possible because  selected tables provide additional labeled examples with expense-type to spend-level relationships that match target month, effectively augmenting the target with useful signals. Discovery and selection therefore address distinct problems, and the improvement holds on two independently collected benchmarks.

\subsection{Ablation studies on \sys and its variants}
\label{sec:ablation}
We now study the effect of hyperparameters $B, k_{max}$ and $|\mathcal{T}|$ on performance of \sys and its variants.

\noindent\textbf{Effect of $B$.}
Figure \ref{fig:ablation_smax} reports profit and models explored as $B$ varies
($k_{max}=B$).
We find that \textit{profit saturates at a value that reflects dataset complexity.} 
On ACSPubCov, \sys plateaus at $B$ = 5 while the surrogates continue improving to $B=7$. On Amazon, $B=5$ recovers virtually all achievable gain, consistent with the selected subset containing 3 compatible product categories. Wilds is the exception: profit continues rising throughout $B$ = 15, showing that the 30-location pool contains complementary locations whose value only emerges at larger initialization sizes. 

\noindent\textit{\underline{Evaluation cost grows super linearly with $B$}.}
On ACSPubCoverage, \sys explores 15 subsets at $B$ = 1 and 409 at $B$ = 7 for 0.37 points of gain beyond $B$ = 3. \sys variants scale  well because of their lightweight computation regardless of $B$, making larger values affordable  even when they are prohibitive for \sys.

\noindent\textit{\underline{Recommended defaults}.} A robust heuristic is to set $B$=$0.2|\mathcal{S}| - 0.5|\mathcal{S}|$
for homogeneous pools.
A higher $B$ is recommended for datasets with high variance in source profits (e.g.\ Wilds, macro-F1 $0.05-0.20$ across locations).

\vspace{1mm}\noindent\textbf{Effect of $k_{max}$.}
Figure \ref{fig:ablation_kmax} shows our observations as $k_{max}$ varies 
and $B$ is fixed.
On ACSPubCov, \sys is flat across all $k_{max}$ while both surrogates improve from $k_{max}=1 \to 3$ (\sysdp 63.03 $\to$ 65.11, \sysgrad 62.67 $\to$ 65.16) and then plateau, showing that a single swap is insufficient when the marginal estimate is approximate. 
Profit varies non monotonically across $k_{max}$ on a single seed (0.234 at $k_{max} = 3$, 0.216 at $k_{max}=9$) on Wilds.
\sys variants remain lightweight regardless of $k_{max}$ since their per-query cost is dominated by surrogate inference rather than model retraining.

\vspace{0.5mm}\noindent\textbf{Effect of $|\mathcal{T}|$.}
Selection is guided by a labeled target sample, whose availability is the requirement with the clearest practical cost.
Holding out a fixed $30\%$ of the target as test, we draw $\mathcal{T}$ at fractions $f \in \{0.05,\dots,0.70\}$ of the remainder, nested so each smaller sample is a subset of a larger subset, and repeat selection at each $f$ (Table~\ref{tab:valsize}).

\begin{table}[h]
  \centering \small
  \setlength{\tabcolsep}{3pt}
  \caption{Effect of validation set size on \sys test profit. $f=$
  fraction of target used for selection; ACSTravelTime MSE
  ($\downarrow$).}
  \label{tab:valsize}
  \begin{tabular}{lccccc}
    \toprule
    $f$ & ACSIncome & ACSPubCov. & ACSTrvlTime & Amazon & Wilds \\
    \midrule
    0.05 & 73.754 & 64.608 & 487.095 & 0.8466 & 0.1822 \\
    0.10 & 73.754 & 64.345 & \textbf{485.206} & 0.8494 & 0.2244 \\
    0.20 & 74.835 & 64.608 & \textbf{485.206} & 0.8484 & \textbf{0.2485} \\
    0.30 & \textbf{75.112} & 64.581 & 485.243 & \textbf{0.8519} & 0.2359 \\
    0.70 & 74.835 & 64.608 & 486.903 & 0.8518 & 0.2337 \\
    \bottomrule
  \end{tabular}
\end{table}

\noindent We observe that \emph{{most benchmarks are insensitive to $|\mathcal{T}|$}.} Profit varies by at most $0.26$ points and $0.005$ F1 across a 14-fold reduction in $|\mathcal{T}|$ (ACSPubCov, Amazon) and by $1.9$ MSE ($0.4\%$) on  ACSTravelTime. A few thousand labeled target examples suffice (with $1,500$ on Amazon).
\emph{Sensitivity arises from optimism in the selection estimate.} ACSIncome loses $1.36$ profit points and Wilds $0.066$ macro-F1 at $f=0.05$ relative to their best setting. On both tasks, 
validation profit minus test profit
decreases monotonically with $|\mathcal{T}|$: $+3.65$ to $-2.02$ on ACSIncome and $+0.085$ to $+0.006$ on Wilds as $f$ goes from $0.05$ to $0.70$, against a magnitude of at most $0.005$ throughout on Amazon. These two benchmarks are the hardest to estimate profit: macro-F1 over $80$ classes from $368$ images, three of which are absent from validation entirely at $f=0.05$, and a $\lambda=50$ parity term that amplifies subgroup estimation error accordingly.  
\sys optimizes  validation estimate directly and is, therefore, most affected when that estimate is optimistic, whereas profile and gradient signals derive from source statistics that a small target draw perturbs less.

\vspace{1mm}\noindent\textbf{Summary.} Ablation studies support the following practical guidelines: across all benchmarks $k_{max}=5$ is a robust default and we did not observe further gains from deeper swaps on any benchmark tested.
\sys variants make larger $k_{max}$ values affordable since their per query cost does not scale with model training time.
$|\mathcal{T}|$ can drop to $5\%$ of target on three of five benchmarks at negligible cost; surrogate variants are preferred not only when evaluation is expensive, but also when target samples are small.

\subsection{Comparison with coreset baselines}
\label{sec:coresets}
A natural alternative to source-level selection is to select individual training points from the union of all candidate sources which, in principle, is more flexible than \sys since it can mix examples from any combination of sources. We compare against three point-level baselines: uniform random sampling, $k$-Center greedy~\cite{sener2017active}, and GradMatch~\cite{killamsetty2021grad} at point granularity, $B=1\%$ of the candidate pool ($B=5\%$ for Wilds, whose $80$ classes need more examples per class for a non-degenerate macro-F1).
The same experiment answers a second question: whether the two granularities \emph{compose}. Alongside \textsc{Cor-all}, which samples from the union of all sources, we run \textsc{Hyb} (rand-\sys), which randomly samples from \sys's selected sources, \textsc{Hyb} (rand-$k$), which samples from $k$ random sources with $k=|S^*|$ and \textsc{Hyb} (best-\sys) that applies the best performing coreset selector for that benchmark to \sys's sources. Table \ref{tab:coresets} reports them over 10 seeds.
 
\begin{table}[t]
  \centering
  \small
  \setlength{\tabcolsep}{3pt}
  \caption{Point-level and hybrid selection, 10 seeds.
  $\dagger$: degenerate classifier.
  Bold: better than \sys.}
  \label{tab:coresets}
  \begin{tabular}{lccccc}
    \toprule
    & ACSInc. & ACSPub. & ACSTrvl$\downarrow$ & Amazon & Wilds \\
    \midrule
    \textsc{Cor-all} (rand)   & 71.281 & 51.168 & 501.617 & 0.8247 & 0.1742 \\
    \textsc{Cor-all} ($k$-C)  & \textbf{74.107} & 51.980 & 489.788 & 0.8340 & 0.2125 \\
    \textsc{Cor-all} (GM)     & 103.480$^\dagger$ & 56.271 & 1566.02 & 0.7755 & 0.1493 \\
    \midrule
    \textsc{Hyb} (rand-$k$)   & 69.091 & 53.015 & 619.643 & 0.8309 & 0.1649 \\
    \textsc{Hyb} (rand-\sys)       & \textbf{73.368} & 64.691 & 488.493 & 0.8444 & 0.1602 \\
    \textsc{Hyb} (best-\sys)       & \textbf{74.890} & 64.691 & \textbf{484.122} & 0.8579 & 0.2034 \\
    \midrule
    \sys                      & 73.253 & 64.818 & 486.026 & 0.8613 & 0.2360 \\
    \bottomrule
  \end{tabular}
\end{table}
 
\noindent\textit{\underline{Point selection cannot avoid whole-source harm}.}
On ACSPublicCoverage \sys beats the best coreset method by $8.5$ points and on
Wilds by $0.024$ macro-F1: these methods draw from a pool of all sources and
cannot escape the influence of harmful ones, whereas \sys removes them from
consideration entirely. GradMatch's $103.48$ on ACSIncome is a degenerate
fairness-gaming solution --- the classifier predicts ``low-income'' for $99\%$
of male and ``high-income'' for $99\%$ of female examples regardless of input,
achieving chance accuracy but near-maximal TPR parity. The cosine proxy first
selects an extreme class-imbalanced coreset ($0.8\%$ positives against $36.3\%$
in the pool), and the fairness-weighted target gradient then drives the model
into that regime; \sys rejects such degeneracies because every swap is scored on
held-out target profit. On ACSTravelTime GradMatch is three times worse than
\sys and worse than random selection, being agnostic to systemic domain shift. $k$-Center is the exception that proves the rule: on Wilds it comes closest to
\sys by spreading points across ResNet-18 feature space, implicitly covering
diverse habitats, but the residual $0.024$ gap is the value of source-level
coherence --- a forest location supplies hundreds of consistent examples of
forest-dwelling species, a structure that uniformly scattered points cannot
reproduce.
 
\noindent\textit{\underline{Source and sample selection are complementary}.}
Restricting pool to \sys's sources recovers $84$--$106\%$ of \sys's advantage over \textsc{Cor-all}, while rand-$k$ control recovers little and is worse than \textsc{Cor-all} on two of  three  tasks; the gain does not come from narrowing the pool, but from selected sources. The combination beats either granularity alone: running $k$-Center with \sys sources gives best ACSIncome result ($74.890$), and restricting the pool repairs GradMatch on Amazon ($0.776\rightarrow0.858$) and ACSTravelTime ($1566\rightarrow808$ MSE), whose misleading gradients come from sources already discarded. In contrast, drawing $5\%$ of rows from all 30 locations ($0.1742$) in Wilds beats drawing $5\%$ from \sys's 14 ($0.1602$), both far below \sys ($0.2360$). A camera-trap location is valuable because it supplies many mutually consistent images of a habitat, and subsampling destroys that structure; at a fixed budget breadth then beats quality, and sources are worth keeping whole. Coreset selection remains competitive where geometric coverage suffices, and composes with \sys as a post-filter over selected sources.

\section{Conclusion and Future Work}
\label{sec:conclusion}
We formalized the problem of selecting sources optimized for a downstream model 
as a target and cost aware combinatorial optimization problem and introduced \sys, a gene-splicing inspired algorithm that iteratively swaps low-value for high-value sources. We additionally provide two surrogate variants \sysdp and \sysgrad that approximate marginal utilities efficiently without per-step retraining.
On \santos-derived benchmarks, \sys-selected subsets exceed the self training upper bound by 1.9-3.1 percentage points of F1, while training on all 19 discovered tables falls 1.1-1.5 points \textit{below} the bound -- selection augments the target with complementary signal that schema-level discovery cannot identify. Across seven benchmarks, \sys improves task utility over the all-sources baseline by 1.8-13.7 points and over greedy selection by up to 11 MSE points and 8.66 macro-F1 points. The surrogates match \sys's quality while reducing model evaluations by 3-7$\times$ at 15 sources, growing upto 77$\times$ at the 50-source ACS pool.
Future work includes (i) feeding selection outcomes back into discovery's
schema-similarity scoring, (ii) multi-target selection, where one subset
serves several downstream tasks, (iii) online selection as new sources
enter the lake, and (iv) approximation guarantees for \sys's local optima
on weakly-submodular landscapes~\cite{bian2017guarantees}.

\begin{acks}
This work was supported by the National Science Foundation (NSF) under Grant $\#2237149$.
\end{acks}
\section*{Artifacts}
\label{sec:artifacts}
Implementation available at \href{https://github.com/am-barish/Splice}{\sys github}.

\balance
\bibliographystyle{ACM-Reference-Format}
\bibliography{ref}

\newpage
\appendix
\section{Theoretical Analysis}
\label{app:theory}

We now show that \sys terminates in a finite number of profit evaluations and returns a solution with a formally characterized optimality property.

\subsection{\sys Convergence}
\label{app:convergence}
We begin with defining $\kappa$-local optimum and use it to establish convergence conditions.
\begin{definition}[$\kappa$-local optimum]
\label{def:klocal}
A subset $S \subseteq \sources$ is a \emph{$\kappa$-local optimum} of $\profit_\target$ if no swap of up to $\kappa$ sources improves profit, i.e. for all $R \subseteq S$ and $D \subseteq \sources \setminus S$ with $|R| = |D| \leq \kappa$:
\begin{equation}
  \profit_\target \bigl((S \setminus R) \cup D\bigr) \leq \profit_\target(S).
  \label{eq:klocal}
\end{equation}
\end{definition}

\begin{theorem}[Convergence]
\label{thm:convergence}
Let $\profit_{\min}$ and $\profit_{\max}$ be finite lower and upper bounds on $\profit_\target$ over $2^\sources$. Introduce a minimum profit improvement $\varepsilon > 0$ required to accept a swap. For fixed $k_{\max}$ and $B$, \sys terminates after at most
\[
O\!\left(B \;\cdot\; k_{\max} \;\cdot\; n \;\cdot\;
\frac{\profit_{\max} - \profit_{\min}}{\varepsilon}\right)
\]
profit evaluations. Furthermore, if \textsc{Splicing} exhaustively evaluates all $k$-swaps (rather than the marginal-gain heuristic), the returned subset is a $k_{\max}$-local optimum of $\profit_\target$ (Definition~\ref{def:klocal}).
\end{theorem}

In our implementation, every strict improvement is accepted; $\varepsilon$ serves as an analytical parameter abstracting the minimum representable positive profit increment on a finite dataset.

\begin{proof}
\textsc{Splicing} accepts a candidate $S'$ only when $\profit_\target(S') \geq \profit_\target(S_{\text{curr}}) + \varepsilon$, so profit increases by at least $\varepsilon$ at each accepted step. Since $\profit_\target$ is bounded in $[\profit_{\min}, \profit_{\max}]$, the number of accepted steps per \textsc{FixedSupport} run is at most $(\profit_{\max} - \profit_{\min})/\varepsilon$.
At each step, \textsc{Splicing} iterates over $k = 1,\dots,k_{\max}$. For each $k$, it computes marginal gains $\gain^-(s)$ for every $s \in S$ ($|S| \leq n$ evaluations) and $\gain^+(u)$ for every $u \in \sources \setminus S$ ($n - |S|$ evaluations), then evaluates the single proposed swap ($1$ evaluation): $O(n)$ evaluations per $k$, and $O(k_{\max} \cdot n)$ per \textsc{Splicing} call. Over all accepted steps and $B$ initializations, total evaluations are bounded by $O\!\left(B \cdot k_{\max} \cdot n \cdot (\profit_{\max} - \profit_{\min})/\varepsilon\right)$.
\sys terminates when no proposed $k$-swap improves profit by $\varepsilon$ for any $k \leq k_{\max}$, i.e. $S_{\text{curr}} = S_{\text{prev}}$ in \textsc{FixedSupport}. If \textsc{Splicing} exhaustively evaluates all $\binom{|S|}{k}\binom{n-|S|}{k}$ candidate $k$-swaps rather than using the marginal-gain heuristic, this termination condition is equivalent to Definition \ref{def:klocal}, and the returned set is a $k_{\max}$-local optimum. The practical algorithm uses marginal gains to guide swap selection, trading the formal local-optimum guarantee for the $O(k_{\max} \cdot n)$ per-step cost.
\end{proof}
\noindent\textbf{Practical implication.}
We observe in experiments (Section~\ref{sec:exp}) that \sys converges in $319-6,385$ evaluations (Figure \ref{fig:method_modex}), far below the exhaustive worst case of $O(n^{2k_{\max}})$, because the profit landscape has few local optima at the operating $k_{\max}\leq B$ and generating the active set with the top-$i$ sources ranked by individual profit already starts the search in high-profit regions of the subset space.

\subsection{Submodularity Ratio of Profit Landscape}
\label{app:approx}

Theorem~\ref{thm:convergence} guarantees termination at a local optimum but says nothing about its distance from the global optimum. We characterize the structure of $\profit_\target$ via the \emph{submodularity ratio} of Das and Kempe \cite{das2011submodular}, used here as a diagnostic.
\begin{definition}[Submodularity ratio \cite{das2011submodular}]
The submodularity ratio of $\profit_\target$ with respect to ground set $U$ {(the universe of items over which $\profit_\target$ is defined)} and parameter $\kappa$ is
\[
\gamma_{U,\kappa} = \min_{\substack{L \subseteq U \\ S \subseteq U \setminus L \\ |S| \le \kappa}}
\frac{\sum_{i \in S}\bigl[\profit_\target(L \cup \{i\}) - \profit_\target(L)\bigr]}
     {\profit_\target(L \cup S) - \profit_\target(L)}.
\]
\end{definition}

{In our setting, $U = \sources$, the full set of candidate sources.}
Das and Kempe's original definition assumes monotone $\profit$, under which $\gamma \in [0,1]$ ($\gamma = 1$ for submodular functions), and show that a greedy solution achieves a $(1-e^{-\gamma})$ approximation for monotone objectives. Our profit function is non monotone because adding a source is not guaranteed to increase utility; therefore, the formal Das-Kempe interpretation does not apply and the approximation guarantee does not transfer to our setting. We, therefore, report the ratio as a qualitative landscape diagnostic rather than a formal bound, with values $>1$ indicating that sum of the individual gains exceeds the combined gain, and values $<0$ indicating that the individual and combined gains disagree. We report the results from our empirical estimation in Section~\ref{app:submod-emp}.

\subsection{\hspace{-2mm}Empirical estimation of submodularity ratio}
\label{app:submod-emp}

As discussed in Section~\ref{app:approx}, for each benchmark, we estimate {$\gamma_{\sources,5}$ -- i.e., the submodularity ratio with ground set equal to the full candidate pool $\sources$ and parameter $\kappa=5$ -- }by sampling 500 random subset pairs $(L, S)$ with $|S| = 5$. For each pair, we compute the ratio of the sum of individual marginal gains to the combined gain. Because the joint gain can be zero or negative in our non-monotone setting (in which case the ratio is undefined or numerically unstable), we filter such pairs and report the median and the fraction of remaining pairs falling into the regimes stated in Section~\ref{app:approx}.

Table~\ref{tab:submod-ratio} reports the results. All benchmarks have a median ratio above 1, indicating that substitution dominates the typical landscape: in 53–78\% of sampled pairs, the sum of individual contributions exceeds the joint contribution, reflecting overlapping signal across sources. Two patterns vary meaningfully across benchmarks. First, the median itself varies by a factor of 2.3$\times$, from a barely substitution flavored ACSTravelTime (median 1.07) to the strongly substitution flavored ACSPublicCoverage (2.37) and IPO (2.43). Second, the opposite sign fraction ranges from 8\% on Amazon to 33\% on ACSTravelTime, characterizing how often individual sources have effects of opposite sign to the joint effect. The fraction of sampled pairs with non-positive joint gain - those filtered before ratio computation - ranges from 12\% on Wilds (where source diversity makes most multi source additions broadly useful) to 47\% on ACSPubCoverage (where source cancellation is most prevalent).

\begin{table}[t]
\centering
\setlength{\tabcolsep}{2.5pt}
\caption{Submodularity ratio diagnostics across benchmarks. \textit{Substitutes} ($\gamma > 1$, sum of individual gains exceeds joint gain), \textit{Partial-complements} ($0 < \gamma \le 1$, joint gain exceeds or matches the sum), \textit{Opposite-sign} ($\gamma < 0$, individual gains and joint gain have opposite signs, only possible in non-monotone settings). \textit{\% Filtered} reports the fraction of all 500 sampled pairs filtered before $\gamma$ computation because the joint gain was non-positive.}
\label{tab:submod-ratio}
\small
\begin{tabular}{lccccc}
\toprule
Benchmark & Median $\gamma$ & \% Sub & \% Part-comp & \% Opp-sign & \% Filtered \\
\midrule
ACSIncome & 1.97 & 77 & 11 & 11 & 26 \\
ACSPubCov & 2.37 & 76 & 11 & 11 & 47 \\
ACSTravelTime & 1.07 & 53 & 14 & 33 & 41 \\
Amazon & 1.71 & 78 & 14 & 8 & 24 \\
IPO & 2.43 & 69 & 10 & 21 & 39 \\
Wilds & 1.44 & 69 & 14 & 17 & 12 \\
\bottomrule
\end{tabular}
\end{table}

These diagnostic patterns track \sys's empirical advantage over greedy forward selection. On benchmarks with high median and low opposite sign fraction (ACSPubCoverage, IPO, Amazon), the landscape rewards local greedy choices, and greedy reaches profit within 0.5–2.2 points of \sys. On ACSTravelTime, the lowest median benchmark with the highest opposite sign fraction, greedy gets stuck in local optima that 1-step additions cannot escape — \sys exceeds greedy by 11 MSE points. On Wilds, the substitution structure is moderate but the per-source value variance is large (single-source macro-F1 spans 0.05–0.20), so multistart initialization and k-swap together are necessary to navigate the landscape — \sys outperforms greedy by 8.66 macro-F1 points, the largest gap in our suite. The submodularity ratio diagnostic is consistent with where \sys's mechanism is most beneficial, even though it does not transfer formal approximation guarantees to the non-monotone setting.

\section{Reference Models for \sysgrad}
\label{app:reference-models}

\sysgrad uses a task-specific reference model trained on the union of all source data before gradient extraction. Architectures are selected to be lightweight (so gradient computation is fast) while remaining expressive enough that gradient directions are informative.

\noindent\textbf{Tabular classification (ACSIncome, ACSPubCoverage, SANTOS).}  
An $\ell_2$-regularized logistic regression model ($\texttt{nn.Linear}(p, 1)$ with sigmoid output), trained with binary cross-entropy for 5 epochs. Gradient dimension: $d_g = p + 1$, where $p$ is the number of input features.

\noindent\textbf{Tabular regression (ACSTravelTime).}  
A single linear layer \\($\texttt{nn.Linear}(p, 1)$), trained with mean squared error for 5 epochs. Gradient dimension: $d_g = p + 1$.

\noindent\textbf{Image classification (Wilds).}  
A linear head $\texttt{nn.Linear}(d_{\text{feat}}, c)$ over frozen ResNet-18 backbone features. The head is warm-started with cross-entropy for 5 epochs on the union of all source data; gradients are then recomputed against a differentiable soft-macro-F1 loss~\cite{eban2017scalable} so that the direction matches the macro-F1 evaluation metric. Gradient dimension: $d_g = d_{\text{feat}} \cdot c + c$ (full weight and bias gradient).

\noindent\textbf{Text classification (Amazon).}  
A two-layer MLP head \\$\texttt{LayerNorm}\rightarrow\texttt{Linear}(d_\mathrm{feat}, h)\rightarrow\texttt{ReLU}\rightarrow\texttt{Dropout}\rightarrow\texttt{Linear}(h,1)$. The head is warm-started in two phases: first with binary cross-entropy on the union of all source data, then fine-tuned with differentiable soft-F1 loss~\cite{eban2017scalable}. We store only the final-layer weight gradient ($d_g = h$) to keep per-query cost low. Target gradients are computed on stratified positive/negative samples and averaged across 5 random seeds to reduce variance~\cite{zhao2015stochastic}.

\section{Paired differences vs \sys}
\label{app:paired-stats}

Table~\ref{tab:paired-stats} reports the closest pairwise comparisons against \sys on identical replicates. We use the paired difference rather than the marginal intervals of Table~\ref{tab:results-by-task} because replicate-to-replicate variance is large and shared across methods: on ACSIncome, for instance, every method's profit varies by about $2.5$ points across the 30 target splits (sd $\approx 0.6$), while the standard deviation of the \sys${-}$\grasp \emph{difference} is $0.22$. Exact ties are common on the small ACS pools, where two searches frequently converge on the same subset.
 
\noindent\textbf{Wilds \textsc{Grasp}.} Table~\ref{tab:results-by-task} reports
\textsc{Grasp} on Wilds without a confidence interval. A single run requires
14,606 model trainings and 12.4 hours; ten seeds would exceed five days of
compute on the evaluation machine, so we report the single-seed value and do
not claim a significance test against it. The paired comparisons in
Table~\ref{tab:paired-stats} accordingly omit \textsc{Grasp} on Wilds.

\begin{table}[t]
\caption{
Paired difference in profit, \sys minus the competing method, on identical target splits/seeds. Positive favours \sys. $\Delta$ is the mean paired difference and the 95\% CI is $\Delta$ $\pm$ 1.96 SE over replicates. $p$ from a two-sided paired $t$-test, Holm-corrected over all comparisons against \sys within each benchmark.
}

\label{tab:paired-stats}
\small
\begin{tabular}{lccrrlrrr}
\toprule
Benchmark & vs. & $\Delta$ (95\% CI) & $p$ \\
\midrule
\multirow{4}{*}{ACSIncome}
 & \textsc{AllSrc}     & $+1.749\ [1.63, 1.87]$   & $<10^{-4}$ \\
 & \textsc{DsDM}       & $+0.333\ [0.15, 0.51]$   & $0.007$ \\
 & \textsc{Grasp}      & $+0.008\ [-0.07, 0.09]$  & $0.836$ \\
 & \textsc{Splice-DP}  & $-0.063\ [-0.16, 0.04]$  & $0.473$ \\
\midrule
\multirow{3}{*}{ACSPubCov}
 & \textsc{AllSrc}     & $+13.690\ [13.65, 13.73]$ & $<10^{-4}$ \\
 & \textsc{Grasp}      & $+0.044\ [0.02, 0.07]$    & $0.005$ \\
 & \textsc{Splice-Grad}& $-0.060\ [-0.08, -0.04]$  & $<10^{-4}$ \\
\midrule
\multirow{3}{*}{ACSTravelTime}
 & \textsc{AllSrc}     & $+14.836\ [14.48, 15.19]$ & $<10^{-4}$ \\
 & \textsc{Grasp}      & $-0.993\ [-1.26, -0.73]$  & $<10^{-4}$ \\
\midrule
\multirow{3}{*}{Amazon}
 & \textsc{AllSrc}     & $+0.0299\ [0.022, 0.037]$ & $<10^{-3}$ \\
 & \textsc{DsDM}       & $+0.0063\ [0.004, 0.009]$ & $0.001$ \\
 & \textsc{Grasp}      & $-0.0001\ [-0.0004, 0.0002]$ & $0.467$ \\
\midrule
\multirow{3}{*}{Wilds}
 & \textsc{AllSrc}     & $+0.0215\ [0.014, 0.029]$ & $0.001$ \\
 & \textsc{DsDM}       & $+0.0197\ [0.009, 0.031]$ & $0.020$ \\
 & \textsc{Splice-DP}  & $+0.0110\ [-0.000, 0.023]$ & $0.180$ \\
\bottomrule
\end{tabular}
\end{table}

 \begin{table}[h]
  \centering \small
  \caption{ACSPubCov source profiles, ordered by solo profit.}
  \label{tab:prof-pubcov}
  \begin{tabular}{llrrrrr}
    \toprule
    State & Sel. & Rows & Mismatch & Shift & Solo & Marginal \\
    \midrule
    CA & & 729{,}432 & 0.211 & 4.21 & 62.23 & $-5.89$ \\
    ME & \checkmark & 23{,}699 & 0.254 & 3.26 & 61.55 & $+0.09$ \\
    ID & \checkmark & 30{,}490 & 0.362 & 3.65 & 59.79 & $+0.53$ \\
    MI & \checkmark & 186{,}896 & 0.230 & 3.01 & 58.05 & $+1.81$ \\
    MD & & 93{,}984 & 0.279 & 3.10 & 57.86 & $-1.04$ \\
    GA & & 194{,}141 & 0.358 & 3.74 & 57.75 & $-5.39$ \\
    FL & & 367{,}768 & 0.337 & 3.81 & 57.44 & $-2.14$ \\
    MN & & 86{,}374 & 0.260 & 3.00 & 55.63 & $-0.14$ \\
    NY & \checkmark & 356{,}554 & 0.183 & 2.38 & 54.18 & $+7.77$ \\
    WI & & 98{,}558 & 0.303 & 1.54 & 53.17 & $-7.40$ \\
    OH & & 217{,}837 & 0.239 & 2.30 & 49.70 & $-2.45$ \\
    NE & & 32{,}572 & 0.383 & 2.80 & 48.12 & $-0.06$ \\
    NC & & 191{,}528 & 0.320 & 2.32 & 45.20 & $-7.88$ \\
    TX & & 502{,}018 & 0.379 & 1.69 & 44.68 & $-17.97$ \\
    UT & & 56{,}819 & 0.430 & 1.84 & 44.52 & $-8.99$ \\
    \midrule
    \multicolumn{2}{l}{\emph{Mean, selected}}   & 149,410 & 0.257 & 3.08 & 58.39 & $+2.55$ \\
    \multicolumn{2}{l}{\emph{Mean, discarded}}  & 233,730 & 0.318 & 2.76 & 52.39 & $-5.40$ \\    
    \bottomrule
  \end{tabular}
\end{table}
\section{Per-source profiles}
\label{app:profiles}
 
Section~\ref{sec: effectiveness} argues that no individual source characteristic consistently indicates whether a source is selected. This appendix gives the evidence one benchmark at a time, each
chosen to isolate one of the three failure modes of Section~\ref{sec:intro}:
source cancellation on ACSPubCov, feature coverage gaps on Wilds, and label
mismatch on Amazon. Throughout, \emph{Shift} is the $\ell_2$ distance between
source and target feature centroids, \emph{Solo} is profit when training on
that source alone, and \emph{Marginal} is the source's contribution to the
subset \sys{} returns. Each table closes with the mean over selected and over
discarded sources, which is the comparison the discussion turns on.
 
\subsection{ACSPublicCoverage: source cancellation}
\label{app:profiles-pubcov}
 
\sys{} keeps 4 of 15 states. The ranking by solo profit and the ranking by
marginal contribution disagree at the top: California has the highest solo
profit of any state (62.23) and a negative marginal contribution ($-5.89$), while New York ranks ninth by solo profit (54.18) and
contributes the most of any source ($+7.77$). Selected states average 58.39
solo against 52.39 for discarded ones, a gap far smaller than the gap in
marginal contribution ($+2.55$ against $-5.40$). Texas shows the mechanism at
its most extreme: the second-largest pool of rows in the benchmark, and a
marginal contribution of $-17.97$. Pooling all fifteen scores 51.12 against
\sys{}'s 64.81 precisely because contributions of this magnitude cancel.
 
\begin{table}[]
\caption{Wilds source profiles. Cls: classes present in the source;
Cov: fraction of target classes covered.}
\label{tab:wilds-profiles}
\small
\begin{tabular}{rrrrrrr}
\toprule
Src & Imgs & Cls & Cov & Shift & Solo & Marginal \\
\midrule
\multicolumn{7}{l}{\emph{Selected by \sys{}}} \\
28 & 3,719 & 17 & 0.452 & 10.06 & 0.0669 & $+0.029$ \\
2  & 4,100 & 24 & 0.548 &  8.14 & 0.0574 & $+0.062$ \\
10 & 3,690 & 22 & 0.516 &  8.47 & 0.0549 & $+0.070$ \\
9  & 2,620 & 23 & 0.516 &  7.92 & 0.0477 & $+0.090$ \\
20 & 2,420 & 15 & 0.419 &  9.31 & 0.0279 & $+0.038$ \\
27 & 2,400 & 19 & 0.065 & 10.88 & 0.0222 & $+0.025$ \\
11 & 3,786 &  4 & 0.097 & 11.42 & 0.0184 & $+0.060$ \\
15 & 3,448 &  4 & 0.097 & 10.73 & 0.0184 & $+0.052$ \\
12 & 2,394 &  6 & 0.129 &  9.85 & 0.0184 & $+0.048$ \\
14 & 7,639 &  6 & 0.129 &  8.96 & 0.0184 & $+0.027$ \\
1  & 2,112 &  3 & 0.065 & 12.31 & 0.0184 & $+0.016$ \\
16 & 2,573 &  2 & 0.065 & 14.54 & 0.0184 & $+0.008$ \\
29 & 2,810 & 20 & 0.452 &  7.63 & 0.0142 & $+0.022$ \\
24 & 8,844 & 15 & 0.065 &  9.70 & 0.0077 & $+0.016$ \\
\midrule
\multicolumn{7}{l}{\emph{Discarded}} \\
4  & 2,802 & 21 & 0.516 &  6.44 & 0.1095 & $-0.051$ \\
3  & 4,190 & 25 & 0.581 &  5.98 & 0.0765 & $-0.067$ \\
13 & 2,219 & 18 & 0.452 &  6.72 & 0.0648 & $-0.033$ \\
26 & 2,260 & 22 & 0.516 &  6.31 & 0.0632 & $-0.028$ \\
7  & 2,500 & 19 & 0.548 &  6.15 & 0.0567 & $-0.064$ \\
0  & 3,640 & 21 & 0.484 &  6.28 & 0.0545 & $-0.068$ \\
17 & 3,982 & 20 & 0.548 &  6.03 & 0.0515 & $-0.022$ \\
19 & 3,327 & 18 & 0.452 &  6.87 & 0.0490 & $-0.014$ \\
22 & 3,038 & 14 & 0.355 &  7.44 & 0.0431 & $-0.009$ \\
23 & 2,220 & 13 & 0.355 &  7.20 & 0.0334 & $-0.034$ \\
18 & 4,528 & 13 & 0.323 &  7.35 & 0.0280 & $+0.002$ \\
25 & 3,770 & 25 & 0.581 &  5.52 & 0.0251 & $-0.022$ \\
6  & 3,290 & 14 & 0.419 &  7.02 & 0.0207 & $-0.069$ \\
5  & 3,069 &  7 & 0.097 &  8.11 & 0.0185 & $-0.080$ \\
21 & 6,855 &  2 & 0.065 &  8.60 & 0.0184 & $-0.014$ \\
8  & 2,840 & 21 & 0.484 &  6.98 & 0.0162 & $-0.039$ \\
\midrule
\multicolumn{3}{l}{\emph{Mean, selected}}  & 0.258 & 9.28 & 0.0293 & $+0.040$ \\
\multicolumn{3}{l}{\emph{Mean, discarded}} & 0.423 & 6.81 & 0.0456 & $-0.038$ \\
\bottomrule
\end{tabular}
\end{table}

\subsection{Wilds: feature coverage gaps}
\label{app:profiles-wilds}
 
Wilds inverts the pattern one would expect. \sys{} keeps 14 of 30 locations,
and the selected ones are \emph{worse} on every per-source statistic: lower
solo profit (0.029 against 0.046), narrower class coverage (0.258 against
0.423), and larger distance from the target (9.28 against 6.81). Six selected
locations share an identical solo macro-F1 of 0.0184 and contain only 2--6
classes each, yet contribute $+0.008$ to $+0.060$. These are the sources that
cover the tail of the label space; the broad-coverage locations that score
well alone sample the same head classes as each other, so their marginal
value is negative once one of them is already active. Location 4 is the clear
case: the highest solo profit in the pool (0.1095), 21 classes, and a
marginal contribution of $-0.051$.

\subsection{Amazon: label mismatch}
\label{app:profiles-amazon}
 
\sys{} keeps 3 of 24 categories. The two highest-solo categories are both
discarded (Industrial \& Scientific at 0.829, Health \& Household at 0.817),
while the selected three rank 3rd, 4th and 14th. Solo profit spans a narrow
0.713--0.829 across the pool, so it separates categories weakly; marginal contributions span $-0.030$ to $+0.036$ and do not track it.
Label mismatch is likewise not a sufficient criterion on its own: Software
and Subscription Boxes have the two largest mismatches (0.127, 0.161) and are
discarded, but so are eighteen categories with mismatch below 0.06.

\begin{table}[htbp]
  \centering \small \setlength{\tabcolsep}{4pt}
  \caption{Amazon source profiles, ordered by solo profit. \emph{Pos.}: positive-class rate.}
  \label{tab:prof-amazon}
  \begin{tabular}{llrrrr}
    \toprule
    Category & Sel. & Pos. & Mismatch & Solo & Marginal \\
    \midrule
    Industrial \& Sci. & & 0.842 & 0.027 & 0.829 & $-0.008$ \\
    Health \& Household & & 0.812 & 0.003 & 0.817 & $-0.028$ \\
    Kindle Store & \checkmark & 0.798 & 0.017 & 0.807 & $+0.036$ \\
    Grocery \& Gourmet & \checkmark & 0.830 & 0.015 & 0.805 & $+0.012$ \\
    Software & & 0.688 & 0.127 & 0.804 & $-0.030$ \\
    Home \& Kitchen & & 0.828 & 0.013 & 0.791 & $-0.016$ \\
    Patio, Lawn \& Garden & & 0.806 & 0.009 & 0.789 & $-0.019$ \\
    Tools \& Home Impr. & & 0.830 & 0.015 & 0.788 & $-0.010$ \\
    Books & & 0.828 & 0.013 & 0.787 & $-0.001$ \\
    Office Products & & 0.870 & 0.055 & 0.784 & $-0.004$ \\
    Arts, Crafts \& Sewing & & 0.866 & 0.051 & 0.784 & $-0.011$ \\
    Electronics & & 0.848 & 0.033 & 0.779 & $-0.025$ \\
    Video Games & & 0.798 & 0.017 & 0.773 & $-0.023$ \\
    Musical Instruments & \checkmark & 0.824 & 0.009 & 0.767 & $+0.010$ \\
    CDs \& Vinyl & & 0.912 & 0.097 & 0.763 & $-0.012$ \\
    Sports \& Outdoors & & 0.848 & 0.033 & 0.762 & $-0.012$ \\
    Baby Products & & 0.832 & 0.017 & 0.759 & $-0.010$ \\
    Automotive & & 0.840 & 0.025 & 0.757 & $-0.004$ \\
    Clothing, Shoes \& Jew. & & 0.856 & 0.041 & 0.757 & $-0.014$ \\
    Appliances & & 0.846 & 0.031 & 0.746 & $-0.003$ \\
    Pet Supplies & & 0.790 & 0.025 & 0.735 & $-0.012$ \\
    Subscription Boxes & & 0.654 & 0.161 & 0.733 & $-0.013$ \\
    Beauty \& Personal Care & & 0.786 & 0.029 & 0.726 & $-0.019$ \\
    Toys \& Games & & 0.872 & 0.057 & 0.713 & $-0.015$ \\
    \midrule
    \multicolumn{2}{l}{\emph{Mean, selected} (3)}   & 0.817 & 0.014 & 0.793 & $+0.020$ \\
    \multicolumn{2}{l}{\emph{Mean, discarded} (21)} & 0.826 & 0.042 & 0.770 & $-0.014$ \\
    \bottomrule
  \end{tabular}
\end{table}

\end{document}